\documentclass[sn-mathphys]{sn-jnl}

\jyear{2022}%

\theoremstyle{thmstyleone}%
\newtheorem{theorem}{Theorem}
\newtheorem{proposition}[theorem]{Proposition}%

\usepackage{booktabs} 
\usepackage{arydshln}

\theoremstyle{thmstyletwo}%

\theoremstyle{thmstylethree}%

\usepackage{amsmath,amssymb} 
\usepackage{cleveref}
\usepackage{color}
\usepackage{multirow}
\usepackage{graphicx}
\usepackage{comment}
\usepackage{changepage}
\usepackage{wrapfig}
\usepackage{colortbl} 
\usepackage{arydshln} 
\usepackage{amsthm}
\usepackage{bm}
\usepackage{rotating}
\usepackage{pdfpages}

\usepackage{subfiles}

\definecolor{darkgreen}{RGB}{5,102,8}

\begin{document}


\title[ ]{Incorporating neuro-inspired adaptability for continual learning in artificial intelligence}


\author[1,2,3]{\fnm{Liyuan} \sur{Wang}}\email{wly19@tsinghua.org.cn}
\equalcont{These authors contributed equally to this work.}

\author[1]{\fnm{Xingxing} \sur{Zhang}}\email{xxzhang1993@gmail.com}
\equalcont{These authors contributed equally to this work.}

\author[2,3]{\fnm{Qian} \sur{Li}}\email{liqian8@tsinghua.edu.cn}

\author[4]{\fnm{Mingtian} \sur{Zhang}}\email{m.zhang@cs.ucl.ac.uk}

\author[1]{\fnm{Hang} \sur{Su}}\email{suhangss@tsinghua.edu.cn}

\author*[1]{\fnm{Jun} \sur{Zhu}}\email{dcszj@tsinghua.edu.cn}

\author*[2,3]{\fnm{Yi} \sur{Zhong}}\email{zhongyithu@tsinghua.edu.cn}

\affil[1]{Dept. of Comp. Sci. \& Tech., Institute for AI, BNRist Center, Tsinghua-Bosch Joint ML Center, THBI Lab, Tsinghua University, Beijing, China}

\affil[2]{School of Life Sciences, IDG/McGovern Institute for Brain Research, Tsinghua University, Beijing, China}

\affil[3]{Tsinghua-Peking Center for Life Sciences, Beijing, China}

\affil[4]{Centre for Artificial Intelligence, UCL, London, UK}


\abstract{
Continual learning aims to empower artificial intelligence (AI) with strong adaptability to the real world. For this purpose, a desirable solution should properly balance memory stability with learning plasticity, and acquire sufficient compatibility to capture the observed distributions. Existing advances mainly focus on preserving memory stability to overcome catastrophic forgetting, but remain difficult to flexibly accommodate incremental changes as biological intelligence (BI) does. By modeling a robust \textit{Drosophila} learning system that actively regulates forgetting with multiple learning modules, here we propose a generic approach that appropriately attenuates old memories in parameter distributions to improve learning plasticity, and accordingly coordinates a multi-learner architecture to ensure solution compatibility. Through extensive theoretical and empirical validation, our approach not only clearly enhances the performance of continual learning, especially over synaptic regularization methods in task-incremental settings, but also potentially advances the understanding of neurological adaptive mechanisms, serving as a novel paradigm to progress AI and BI together.
}




\maketitle

\section{Introduction}\label{sec1}

Continual learning, also known as lifelong learning, provides the foundation for artificial intelligence (AI) systems to accommodate real-world changes.
Since the external environment tends to be highly dynamic and unpredictable, an intelligent agent needs to learn and remember throughout its lifetime \cite{chen2018lifelong,parisi2019continual,kudithipudi2022biological}. Numerous efforts have been devoted to preserving memory stability to mitigate catastrophic forgetting in artificial neural networks, where parameter changes for learning each new task well typically result in a dramatic performance drop of the old tasks \cite{mccloskey1989catastrophic,mcclelland1995there,wang2023comprehensive}.
Representative strategies include selectively stabilizing parameters \cite{kirkpatrick2017overcoming,aljundi2018memory,zenke2017continual,chaudhry2018riemannian,ritter2018online}, recovering old data distributions \cite{rebuffi2017icarl,shin2017continual,wang2021memory} and allocating dedicated parameter subspaces \cite{serra2018overcoming,fernando2017pathnet}, etc. 
However, they usually achieve only modest improvements in specific scenarios, with effectiveness varying widely across experimental settings (e.g., task type and similarity, input size, number of training samples, etc.) \cite{parisi2019continual,delange2021continual,kudithipudi2022biological}.
As a result, there is still a huge gap between existing advances and realistic applications.

To overcome this limitation, we theoretically analyze the key factors on which continual learning performance depends, suggesting a broader objective beyond the current focus. Specifically, in order to perform well on all tasks ever seen, a desirable solution should properly balance memory stability of old tasks with learning plasticity of new tasks, while being adequately compatible to capture their distributions (see Methods for details). For example, if you want to accommodate a sequence of cakes (i.e., incremental tasks) into a bag (i.e., a solution), you should optimize the efficiency of space allocation for each cake as well as the total space of the bag, rather than simply freezing the old cakes.

Since biological learning systems are natural continual learners that exhibit strong adaptability to real-world changes \cite{kudithipudi2022biological,hadsell2020embracing,parisi2019continual}, we argue that they have been equipped with effective strategies to address the above challenges.
In particular, the $\gamma$ subset of the \textit{Drosophila} mushroom body ($\gamma$MB) is a biological learning system that is essential for coping with different tasks in succession and enjoys relatively clear and in-depth understanding at both functional and anatomical levels \cite{shuai2010forgetting,cohn2015coordinated,waddell2016neural,modi2020drosophila,aso2014mushroom,aso2016dopaminergic,gao2019genetic,zhao2023genetic} (Fig.~1a), which emerges as an excellent source for inspiring continual learning in AI.

As a key functional advantage, the $\gamma$MB system can regulate memories in distinct ways to optimize memory-guided behaviors in changing environments \cite{richards2017persistence,shuai2010forgetting,dong2016inability,zhang2018active,davis2017biology,gao2019genetic,mo2022age}. First, old memories are actively protected from new disruption by strengthening the previously-learned synaptic changes \cite{zhang2018active,mo2022age}. This idea of selectively stabilizing parameters has been widely used to alleviate catastrophic forgetting in continual learning \cite{kirkpatrick2017overcoming,aljundi2018memory,zenke2017continual,chaudhry2018riemannian}. Besides, old memories can be actively forgotten for better adapting to a new memory \cite{shuai2010forgetting,dong2016inability,gao2019genetic,cervantes2016scribble}. There are specialized molecular signals to regulate the speed of memory decay \cite{davis2017biology,noyes2021memory}, whose activation reduces the persistence of outdated information, while inhibition exhibits the opposite effect \cite{shuai2010forgetting,dong2016inability,gao2019genetic,cervantes2016scribble}. 
However, the benefits of active forgetting for continual learning remain to be explored \cite{kudithipudi2022biological}. Here, we propose a functional strategy that incorporates active forgetting together with stability protection for a better trade-off between new and old tasks, where the active forgetting part is formulated as appropriately attenuating old memories in parameter distributions and optimized by a synaptic expansion-renormalization process. Without compromising old tasks, our proposal can greatly enhance the performance of new tasks by eliminating the past conflicting information. 

We further explore the organizing principles of the $\gamma$MB system that support its function, which employs five compartments ($\gamma$1-5) with dynamic modulations to perform continual learning in parallel \cite{cohn2015coordinated,waddell2016neural,modi2020drosophila,aso2014mushroom,aso2016dopaminergic}. As shown in Fig.~1a, the sensory information is incrementally input from Kenyon cells (KCs), while the valence is conveyed by dopaminergic neurons (DANs) \cite{waddell2016neural,cognigni2018right,amin2019neuronal}. The outputs of these compartments are carried by distinct MB output neurons (MBONs) and integrated in a weighted-sum fashion to guide adaptive behaviors \cite{cognigni2018right,amin2019neuronal,modi2020drosophila}. 
In particular, the DANs allow for distinct learning rules and forgetting rates in each compartment, where the latter has been shown important for processing sequential conflicting experiences \cite{aso2016dopaminergic,handler2019distinct,mccurdy2021dopaminergic,berry2012dopamine,berry2018dopamine}.
Inspired by this, we design a specialized architecture of multiple parallel learning modules, which can ensure solution compatibility for incremental changes by coordinating the diversity of learners' expertise.
Interestingly, adaptive implementations of the proposed functional strategy can naturally serve this purpose through adjusting the target distribution of each learner, suggesting that the neurological adaptive mechanisms are highly synergistic rather than operating in isolation.

Through satisfying the identified criteria, our approach exhibits superior generality across various continual learning benchmarks and achieves significant performance gains.
We further cross-validate the computational model with biological findings, so as to better understand the underpinnings of real-world adaptability for both AI and BI.

\section{Result}\label{sec2}
\subsection{Active Forgetting with Stability Protection}\label{sec2.1}

A central challenge of continual learning is to resolve the mutual interference between new and old tasks due to their distribution differences.
The functional advantages of the $\gamma$MB system suggest that stability protection and active forgetting are both important \cite{richards2017persistence,shuai2010forgetting,dong2016inability,zhang2018active,davis2017biology,gao2019genetic,mo2022age} (Fig.~1b), although current efforts mainly focus on the former to prevent catastrophic forgetting \cite{mccloskey1989catastrophic,mcclelland1995there}. 
Here we formulate this process with the framework of Bayesian learning, which has been hypothesized to well model biological synaptic plasticity by tracking the probability distribution of synaptic weights under dynamic sensory inputs \cite{aitchison2021synaptic,schug2021presynaptic}. 
We briefly describe a simple case of two tasks (Fig.~2a) and leave the full details to Methods.

Let's consider a neural network with parameters $\theta$ continually learning tasks $A$ and $B$ from their training data \(D_A\) and \(D_B\) in order to perform well on their test data, which is called a ``continual learner''. From a Bayesian perspective, the learner first places a prior distribution $p(\theta)$ on $\theta$. After learning task $A$, the learner updates the belief of the parameters, resulting in a posterior distribution $p(\theta\vert D_A)\propto p(D_A \vert \theta)p(\theta)$ that incorporates the knowledge of task $A$. Then, task $A$ can be performed successfully by finding a mode of the posterior: $\theta_{A}^* = \arg\max_\theta
\log \, p(\theta \vert D_A) $. For learning task $B$, $p(\theta \vert D_A)$ becomes the prior and the posterior $p(\theta \vert D_A, D_B)\propto p(D_B \vert \theta)p(\theta \vert D_A)$ will further incorporate the knowledge of task $B$. Similarly, the learner needs to find $\theta_{A,B}^* = \arg\max_\theta
\log \, p(\theta \vert D_A, D_B)$, corresponding to maximizing both $\log p(D_B\vert \theta)$ for learning task $B$ and $\log p(\theta \vert D_A)$ for remembering task $A$.

However, due to the differences in data distribution, remembering old tasks precisely can increase the difficulty of learning each new task well.
Inspired by the biological active forgetting, we introduce a forgetting rate $\beta$ and replace $p(\theta \vert D_A)$ with 
\begin{small}
\begin{equation}
\hat{p}(\theta \vert D_A, \beta) = \frac{p(\theta \vert D_A)^{(1 - \beta) } p(\theta)^{\beta }} {Z},
\label{eq:New_Posterior}
\end{equation}
\end{small}
where $p(\theta)$ is a non-informative prior without incorporating old knowledge \cite{wang2021afec}. $Z$ is a $\beta$-dependent normalizer that keeps $\hat{p}$ a normalized probability distribution (Supplementary Section~1.1). $\hat{p}$ tends to forget task \(A\) when $\beta\rightarrow 1$, while be dominated by $p(\theta \vert D_A)$ with full old knowledge when $\beta\rightarrow 0$. For the new target $p(\theta \vert D_A, D_B, \beta) \propto p(D_B \vert \theta)\hat{p}(\theta \vert D_A, \beta)$, we derive the loss function 
\begin{small}
\begin{equation}
\begin{split}
\mathcal{L}_{\rm{Reg}}^{\rm{AF}}(\theta) =  
&\mathcal{L}_{B}(\theta)  + \underbrace{ \frac{\lambda_{\rm{SP}} }{2}\sum_m F_{A,m} (\theta_m - \theta_{A,m}^*)^2}_{\rm{Stability \,\, Protection}} + \underbrace{\frac{\lambda_{\rm{AF}}}{2}\sum_m I_{e,m} (\theta_m - \theta_{e,m})^2}_{\rm{Active \,\, Forgetting}}. \\
\end{split}
\label{eq:AFEC_objective} 
\end{equation}
\end{small}
$\mathcal{L}_{B}(\theta) $ is the loss function of learning task $B$. $\lambda_{\rm{SP}}$ and $\lambda_{\rm{AF}}$ are hyperparameters that control the strengths of two regularizers responsible for stability protection and active forgetting, respectively. 
The stability protection part is to selectively penalize the deviance of each parameter $\theta_m$ from $\theta_{A,m}^*$ depending on its ``importance'' for old task(s), estimated by the Fisher information $F_{A,m}$. 

The optimization of active forgetting can be achieved in two equivalent ways, i.e., AF-1 and AF-2 (Fig.~2b). 
They both encourage the network parameters $\theta$ to renormalize with an ``expanded'' set of parameters $\theta_e$ when learning task $B$. 
For AF-1, $\theta_{e,m} = 0$ is ``empty'' with equal selectivity $I_{e,m}=1$ for renormalization, where the active forgetting term becomes the $L_2$ norm of $\theta$. 
The hyperparameters $\lambda_{\rm{AF}} \propto \beta$ and $\lambda_{\rm{SP}} \propto  1-\beta$, indicating that the old memories are directly affected.
For AF-2, $\theta_{e,m}=\theta_{B,m}^*$ is the optimal solution for task $B$ only, obtained from optimizing $\mathcal{L}_{\rm{B}}(\theta_e) $, and $I_{e,m}=F_{B,m}$ is the Fisher information. The forgetting rate is fully integrated into $\lambda_{\rm{AF}} \propto  \beta / (1-\beta)$ and is independent of $\lambda_{\rm{SP}}$. 
In the absence of active forgetting ($\beta = 0$), the loss function in Eq.~\eqref{eq:AFEC_objective} is left with only $\mathcal{L}_{\rm{B}}(\theta)$ and the stability protection term, which is (approximately \cite{benzing2022unifying}) equivalent to regular synaptic regularization methods such as EWC \cite{kirkpatrick2017overcoming}. In particular, since the loss functions of these methods \cite{kirkpatrick2017overcoming,aljundi2018memory,zenke2017continual,chaudhry2018riemannian} typically have a similar form and differ only in the metric for estimating the parameter importance \cite{benzing2022unifying} (Methods Eq.~\eqref{eq:Regularization}), our proposal can be naturally combined with them by plugging in the active forgetting term.

For biological neural networks, active forgetting is able to remove outdated information and provide flexibility for adapting to a new memory \cite{shuai2010forgetting,dong2016inability,richards2017persistence}. 
This strategy is essential for \textit{Drosophila} to cope with the interference of previous tasks \cite{shuai2010forgetting,dong2016inability,bouton1993context,zhang2018active}. 
Here we theoretically analyze how this benefit is achieved in our computational model. First, an appropriate forgetting rate $\beta$ is able to improve the probability of learning new tasks well through attenuating old memories in $\theta$ (Methods Eq.~\eqref{eq:Beta_Flexibility}), which can be empirically determined by a grid search of $\lambda_{\rm{AF}}$ and/or $\lambda_{\rm{SP}}$. 
Second, when $\theta$ moves to the neighborhood of an empirical optimal solution, the active forgetting term in Eq.~\eqref{eq:AFEC_objective} can minimize the upper bound of generalization errors for continual learning, especially for new tasks (Methods Proposition~\ref{AF}).

Now we evaluate the efficacy of active forgetting on three continual learning benchmarks for visual classification tasks. They are all constructed from the CIFAR-100 dataset \cite{krizhevsky2009learning} of 100-class colored images but with different degrees of overall knowledge transfer \cite{wang2021afec}. 
As shown in Fig.~2c, the proposed active forgetting can largely enhance the average accuracy of all tasks, using EWC \cite{kirkpatrick2017overcoming} as a baseline for preserving memory stability. 
Then we analyze the benefits of active forgetting on learning plasticity and memory stability with the metrics of forward transfer and backward transfer, respectively, where the former is clearly dominant. 
Similar results are observed when plugging the active forgetting term in other synaptic regularization methods that only preserve memory stability (Supplementary Fig.~1). In contrast, active forgetting fails to improve the joint training performance (Supplementary Fig.~4a), suggesting that its benefits are specific to continual learning. From visual interpretation of the latest task predictions in Fig.~2d, active forgetting can indeed eliminate the past conflicting information, leading to better recognition of the object itself.

\subsection{Coordination of Multiple Continual Learners}\label{sec2.2}
After demonstrating the benefits of active forgetting together with stability protection for a single continual learner (SCL), we turn to investigate the organizing principles of the $\gamma$MB system where new memory forms and active forgetting happens \cite{shuai2010forgetting,cohn2015coordinated,waddell2016neural,modi2020drosophila,gao2019genetic,berry2012dopamine,berry2018dopamine,handler2019distinct,shuai2015dissecting}. 
Specifically, there are five compartments that process sequential experiences in parallel. The outputs of these compartments are integrated in a weighted sum fashion to guide adaptive behaviors. Inspired by this, we design a $\gamma$MB-like architecture consisting of multiple parallel continual learners (Fig.~3a). Each learner employs a parameter space to learn all tasks, but its dedicated output head is removed and the weighted sum of the previous layer's output is fed into a shared output head, where the output weights of each learner are incrementally updated.

In such a $\gamma$MB-like architecture, the relationship between learners is critical to the performance of continual learning. 
When the diversity of their expertise is properly coordinated, the obtained solution can provide a high degree of compatibility with both new and old tasks. 
Here we present a conceptual illustration via performing tasks $A$, $B$ and $C$ with different similarities (Fig.~3c). Since it is difficult to find a shared optimal solution for all tasks, the SCL has to converge to a high error region for tasks $A$ and $C$. In contrast, the multiple continual learners (MCL) with appropriate diversity allow for division of labor to address task discrepancy and complement their functions as parameter changes. Then, the output weights can integrate respective expertise of each learner into the final prediction.

In general, each learner's expertise is directly modulated by its target distribution, where the proposed functional strategy can naturally serve this purpose. 
With the formulation of active forgetting, the target distribution $p(\theta \vert D_A, D_B, \beta) \propto p(D_B \vert \theta)\hat{p}(\theta \vert D_A, \beta)$ tends to be different for learners with different forgetting rates $\beta$, and vice versa (Fig.~3b). Therefore, we implement the forgetting rates adaptively for these learners to coordinate their relationship. Since the target distribution also depends on $p(D_B\vert \theta)$, we further propose a supplementary modulation that constrains explicitly the differences in predictions between learners, corresponding to adjusting the learning rules for each new task.
We refer to the $\gamma$MB-like architecture with these two modulations as Collaborative continual learners with Active Forgetting (CAF), and provide a formal definition in Methods Eq.~\eqref{eq:CAF_loss}.

For multiple learners with identical network architectures and similar forms of learning objectives, the priority is to obtain adequate differentiation of their expertise. In this case, the forgetting rates serve to diversify these learners, similar to the neurological strategy of decaying old memories differentially in each compartment \cite{aso2016dopaminergic,handler2019distinct,mccurdy2021dopaminergic,berry2012dopamine,berry2018dopamine}. 
In practice, the differences of learners can also arise from their innate randomness, such as the use of dropout and different random initializations, leading to sub-optimal solutions with moderate performance.
This potentially corresponds to the anatomical randomness of KCs receiving olfactory signals in \textit{Drosophila} \cite{chen2023ai,caron2013random,endo2020synthesis}.
At this point, the modulations of forgetting rates and learning rules can provide finer adjustments, e.g., by constraining excessive differences.

Given the same network width for each learner, using more learners (i.e., more parameters) generally results in better performance. However, there is an intuitive trade-off between learner number and width under a limited parameter budget. 
We verify that this trade-off is independent of training data distributions (Supplementary Section~1.3) and is relatively insensitive over a wide range (Supplementary Table~5). 
Therefore, we simply choose five learners ($K=5$) corresponding to the five biological compartments, which employs approximately 5 $\times$ parameters, and then reduce the network width accordingly to keep the total amount of parameters similar to that of the SCL. 
To evaluate the effect of innate diversity, we construct a low-diversity background by removing the dropout and using the same random initialization for each learner, and a high-diversity background by maintaining these randomness factors, where the overall diversity of expertise is evaluated by the average cosine or Euclidean distance between learners' predictions.

As shown in Fig.~3e,f, adaptive implementations of either active forgetting (AF-1) or its supplementary modulation (AF-S) can greatly enhance the performance of MCL, where the degree of improvement varies with the effectiveness of increasing inadequate diversity or reducing excessive diversity in the two backgrounds, respectively. 
In response to different degrees of innate diversity, the respective advantages of AF-1 and AF-S are combined to achieve consistently better performance.
Such modulations enable the multiple learners to effectively divide and cooperate in continual learning (Fig.~3d, Supplementary Fig.~4b,c, Supplementary Fig.~5). They exhibit a clear diversity of task expertise with several experts collaborating on each task, validating the conceptual model in Fig.~3c. 
Accordingly, the performance of CAF is largely superior to that of the SCL, averaging the predictions of five independently-trained continual learners, or using a separate learner for each task (Supplementary Fig.~6a,b). In particular, CAF can improve the SCL by a similar magnitude under different parameter budgets, indicating its outstanding scalability (Supplementary Fig.~7a, Supplementary Fig.~6c).

According to our theoretical analysis in Methods Proposition~\ref{CAF}, the performance of a shared solution for new and old tasks depends on the discrepancy of task distributions and the flatness of loss landscape around it. With respect to these two aspects, we delve more deeply into the benefits of our approach. As for the former, we evaluate the discrepancy of task distributions in feature space via the difficulty of distinguishing them after continual learning \cite{long2015learning,wang2022coscl} (Fig.~4a-c). The large increase in discrimination error suggests that our approach can successfully reconcile this discrepancy.
As for the latter, the solution obtained by ours enjoys a clearly flatter loss landscape (Fig.~4d-f), indicating that it is more robust to modest parameter changes in response to dynamic data distributions. 
Therefore, CAF can update parameters more flexibly than the SCL (Fig.~4g-i), with the performance of new and old tasks simultaneously improved (Supplementary Fig.~7b,c).

Finally, we evaluate CAF under the setting of task-incremental learning \cite{van2022three} and compare it to a range of representative methods \cite{kirkpatrick2017overcoming, aljundi2018memory, zenke2017continual, riemer2018learning, schwarz2018progress, jung2020continual, cha2020cpr}.
We first consider visual classification tasks with different particular challenges. 
Besides the overall knowledge transfer, we employ additionally four benchmark datasets such as Omniglot \cite{lake2015human} for long task sequence with imbalanced class numbers, CUB-200-2011 \cite{wah2011caltech} and Tiny-ImageNet \cite{delange2021continual} for larger scale images, and CORe50 \cite{lomonaco2017core50} for smoothly-changed observations.
As shown in Fig.~5, the performance of all baselines varies widely across experimental settings, while CAF achieves consistently the strongest performance in a plug-and-play manner.
We further experiment with Atari reinforcement tasks, where an agent incrementally learns to play several Atari games (Fig.~6a). The overall performance is evaluated by the normalized accumulated reward (NAR) \cite{jung2020continual,wang2021afec,cha2020cpr}, where the rewards obtained for all tasks ever seen are normalized with the maximum reward of fine-tuning on each task, and then accumulated. Likewise, CAF can greatly enhance the performance of baseline approaches (Fig.~6b,c) through improving both learning plasticity and memory stability (Fig.~6d,e).

\section{Discussion}\label{sec12}

Whether for animals, robots or other intelligent agents, the ability of continual learning is critical for successfully adapting to the real world.
In this work, we draw inspirations from the adaptive mechanisms equipped in a robust biological learning system, and present a generic approach for continual learning in artificial neural networks. Our preliminary versions of some individual components have been presented at top conferences in artificial intelligence \cite{wang2021afec,wang2022coscl}, while the current version enjoys substantial extensions in terms of technical robustness, synergistic cooperation and biological plausibility (Supplementary Section~3).
The superior performance and generality of our approach can facilitate realistic applications, such as smartphone, robotics and autonomous driving, to flexibly accommodate user needs and environmental changes. Meanwhile, the deployment of continual learning avoids retraining all previous data each time the model is updated, which provides an energy-efficient and eco-friendly path for developing AI systems.

To bridge the gap between AI and BI, we carefully avoid involving specific implementations or overly strong assumptions in both theoretical analysis and computational modeling. This consideration not only allows for an adequate exploitation of biological advantages, but also facilitates the emergence of interdisciplinary insights. 
Computationally, our approach is proven to satisfy the key factors on which continual learning performance depends, such as stability, plasticity and compatibility, with active forgetting playing an important role.
This potentially extends the previous focus of preventing catastrophic forgetting in continual learning. Starting with this idea, below we discuss more broadly the connections between AI and BI in adaptability.

In a biological sense, active forgetting allows flexibility to accommodate external changes by removing outdated information \cite{richards2017persistence,ryan2022forgetting,davis2017biology}. 
This perspective is well supported by extensive theoretical and empirical evidence in our computational model.
Recent work in neurobiology is deeply dissecting its underlying mechanisms from molecular to synaptic structural levels, where activation of the molecular signaling that mediates active forgetting initially leads to rapid growth of synaptic structures, but prolonged activation instead leads to their shrinkage \cite{shuai2010forgetting,davis2017biology,ryan2022forgetting,luo1996differential,tashiro2000regulation,hayashi2010disrupted,hayashi2015labelling,wang2021afec}. In our computational model, active forgetting of old memories in parameter distributions can derive two equivalent synaptic expansion-renormalization processes.
These two processes and their linear combinations cover a wide range of possible forms, including whether old memories are directly affected and whether expanded parameters encode new memories, which can serve as testable hypotheses for further research. 
As for the five compartments of the $\gamma$MB system, the modulated forgetting rates have been shown to be important for coping with conflicting memories in succession \cite{aso2016dopaminergic,handler2019distinct,mccurdy2021dopaminergic,berry2012dopamine,berry2018dopamine}. Correspondingly, adaptive implementations of active forgetting help the $\gamma$MB-like architecture to better accommodate incremental changes. Besides, we identify the necessity of regularizing learners' predictions of new tasks, suggesting that the adaptation of learning rules may also contribute to continual learning in a more general context. 

AI and BI share the common goal of adaptation and survival in the real world. These two fields have great potential to inspire each other and progress together. This requires generalized theories and methodologies to integrate their advances, as suggested by our work in continual learning. Subsequent work could further explore the ``natural algorithms'' responsible for other advantages of the biological brain, thereby evolving progressively the current AI systems.

\section{Methods}\label{sec11}

\subsection{Synaptic Expansion-Renormalization\label{method_sr}}
For the case of two tasks, the learner needs to find a mode of the posterior distribution that incorporates the knowledge of tasks $A$ and $B$:
\begin{small}
\begin{align}
\begin{split}
\theta_{A,B}^*&= \arg\max_\theta \log \, p(\theta \vert D_A, D_B) \\
&= \arg\max_\theta\log \, p(D_B\vert \theta) + \log \, p(\theta \vert D_A) - \underbrace{\log \, p(D_B)}_{const.},
\end{split}
\label{eq:Bayesian_CL}
\end{align}
\end{small}
\noindent
where $p(D_B\vert \theta)$ is the loss for task $B$.
Although $p(\theta \vert D_A)$ is generally intractable, we can locally approximate it with a second-order Taylor expansion around $\theta_A^* =\arg\max_\theta \log p(\theta\vert D_A)$, resulting in a Gaussian distribution whose mean is $\theta_A^*$ and precision matrix is the Hessian of the negative log posterior \cite{kirkpatrick2017overcoming,ritter2018online,martens2015optimizing}. To simplify the computation, the Hessian is approximated by the diagonal of the Fisher information matrix: 
\begin{small}
\begin{equation}
F_A = \mathbb{E}[ (\frac{\partial \log p(\theta \vert D_A)}{\partial \theta})(\frac{\partial \log p(\theta \vert D_A)}{\partial \theta})^\top \vert_{\theta_A^*}].
\label{eq:FIM_calculate}
\end{equation}
\end{small}

To improve learning plasticity, we introduce a forgetting rate $\beta$, and replace $p(\theta \vert D_A)$ in Eq.~\eqref{eq:Bayesian_CL} with $\hat{p}(\theta \vert D_A, \beta) = \frac{p(\theta \vert D_A)^{(1 - \beta) } p(\theta)^{\beta }}{Z} $ as Eq.~\eqref{eq:New_Posterior}, where $\beta \in [0,1]$ is deterministic and $Z$ is a $\beta$-dependent normalizer.
Correspondingly, the target distribution $p(\theta \vert D_A, D_B)$ becomes $p(\theta \vert D_A, D_B, \beta)$. 
\(\hat{p}\) has a nice property that it follows a Gaussian distribution if $p(\theta \vert D_A)$ and $p(\theta)$ are both Gaussian (Supplementary Section~1.1), so we can compute $\hat{p}$ in a similar way as we compute $p(\theta \vert D_A)$. A certain value of $\beta$ can maximize the probability of learning each new task well through forgetting the old knowledge, validating the motivation of introducing the non-informative prior in Eq.~\eqref{eq:New_Posterior}:
\begin{small}
\begin{align}
\beta^* &=\arg\max_{\beta} p(D_B \vert D_A,\beta) = \arg\max_{\beta} \int p(D_B \vert \theta)\hat{p}(\theta \vert D_A,\beta)d\theta.
\label{eq:Beta_Flexibility}
\end{align}
\end{small}

With the implementation of active forgetting, the learner needs to find
\begin{small}
\begin{align}
\begin{split}
\theta_{A,B}^*&= \arg\max_\theta \log \, p(\theta \vert D_A, D_B, \beta) \\
& = \arg\max_\theta \, \log \, p(D_B\vert \theta) + \log \, \hat{p}(\theta \vert D_A, \beta) - \underbrace{\log \, p(D_B)}_{const.}      \\
& \overset{\rm{AF}-1}{=} \arg\max_\theta \, \log \, p(D_B\vert \theta) + (1-\beta) \log \, p(\theta \vert D_A) + \beta \log \, p(\theta)   \\
&\overset{\rm{AF}-2}{=}  \arg\max_\theta \, (1-\beta) \log \, p(D_B \vert \theta) + (1-\beta) \log \, p(\theta \vert D_A)  + \, \beta \log \, p(\theta \vert D_B),  \\
\end{split}
\label{eq:New_Objective_Derivation}
\end{align}
\end{small} 

\noindent
which can be optimized in two equivalent ways, i.e., AF-1 and AF-2. 
Correspondingly, we derive the loss function in Eq.~\eqref{eq:AFEC_objective}.

For continual learning of more than two tasks, e.g., $t$ tasks for any $t>2$, the learner needs to find
\begin{small}
\begin{align}
\theta_{1:t}^* = \arg\max_\theta\log \, p(D_t\vert \theta) + \log \, p(\theta \vert D_{1:t-1}, \beta_{1:t-1}) - \underbrace{\log \, p(D_t)}_{const.},
\label{eq:Bayesian_CL_t_Tasks}
\end{align}
\end{small}
\noindent
where $D_{1:t-1} = \bigcup_{i=1}^{t-1} D_{i}$ denotes the training data of previous task(s) and $\beta_{1:t-1} = \{ \beta_i \}_{i=1}^{t-1}$ denotes the previously-used forgetting rate(s). Similarly, we replace the posterior $p(\theta \vert D_{1:t-1}, \beta_{1:t-1})$ that absorbs all information of \(D_{1:t-1}\) with
\begin{small}
\begin{equation}
\hat{p}(\theta \vert D_{1:t-1}, \beta_{1:t-1}, \beta_t)  = \frac{p(\theta \vert D_{1:t-1}, \beta_{1:t-1})^{(1 - \beta_t) } p(\theta)^{\beta_t}}{Z_t},
\label{eq:mixture_posterior_Task_t}
\end{equation}
\end{small}
\noindent
where $Z_t$ is a $\beta_t$-dependent normalizer that keeps $\hat{p}$ a normalized probability distribution.
To simplify the hyperparameter tuning, we adopt an identical forgetting rate in continual learning, i.e., $\beta_i = \beta$ for $i=1,...,t$. Then we obtain the loss function:
\begin{small}
\begin{equation}
\begin{split}
\mathcal{L}_{\rm{Reg}}^{\rm{AF}}(\theta) =  
&\mathcal{L}_{t}(\theta)  + \underbrace{ \frac{\lambda_{\rm{SP}}}{2}\sum_m F_{1:t-1,m} (\theta_m - \theta_{1:t-1,m}^*)^2}_{\rm{Stability \,\, Protection}} + \underbrace{\frac{\lambda_{\rm{AF}}}{2}\sum_m I_{e,m} (\theta_m - \theta_{e,m})^2}_{\rm{Active \,\, Forgetting}}.  \\
\end{split}
\label{eq:AFEC_Objective_t_Task} 
\end{equation}
\end{small}
\noindent
$\theta_{1:t-1}^*$ is the obtained solution for previous tasks, i.e., the old network parameters. For AF-1, $\theta_{e,m}=0$, $I_{e,m}=1$, $\lambda_{\rm{AF}} \propto  \beta$ and $\lambda_{\rm{SP}} \propto  (1-\beta)$. For AF-2, $\theta_{e,m}=\theta_{t,m}^*$, $I_{e,m}=F_{t,m}$,  $\lambda_{\rm{AF}} \propto  \beta / (1-\beta)$ and $\lambda_{\rm{SP}} \propto 1$. $F_{1:t-1}$ is recursively updated by
\begin{small}
\begin{equation}  
{F}_{1:t-1} = {F}_{1:t-2} + {F}_{t-1}. 
\label{eq:FIM_update}
\end{equation} 
\end{small}

When $\beta=0$, the loss function in Eq.~\eqref{eq:AFEC_Objective_t_Task} degenerates to a similar form as regular synaptic regularization methods that only preserve memory stability \cite{kirkpatrick2017overcoming,aljundi2018memory,zenke2017continual,chaudhry2018riemannian}:
\begin{small}
\begin{equation}
\mathcal{L}_{\rm{Reg}}(\theta) =  \mathcal{L}_{t}(\theta)  + \underbrace{ \frac{\lambda_{\rm{SP}}}{2}\sum_m \xi_{1:t-1,m} (\theta_m - \theta_{1:t-1,m}^*)^2}_{\rm{Stability \,\, Protection}}.  \\
\label{eq:Regularization} 
\end{equation}
\end{small}

Since these methods differ mainly in the metric $\xi_{1:t-1}$ of estimating the importance of parameters for performing old tasks \cite{benzing2022unifying}, the active forgetting term can be naturally combined with them. We discuss in more depth the motivation and implementation of active forgetting in Supplementary Section~1.2, including the choice of an appropriate $\beta$, the connections of two equivalent versions, and the technical details of derivation.

\subsection{Multiple Parallel Continual Learners\label{method_mcl}}
The $\gamma$MB-like architecture of multiple continual learners (MCL) adopts $K$ identically structured neural networks $f_{\phi_i}( \cdot ), i=1,...,K$, corresponding to $K$ continual learners ${\rm{L}}_i, i=1,...,K$ with their own parameter spaces.
We remove the dedicated output head of each learner, and feed the weighted sum of the previous layer's output into a shared output head $h_{\varphi}(\cdot)$ to make predictions, where the output weight of each learner $g_i, i=1,...,K$ is updated incrementally. Then, the final prediction becomes $p(\cdot) =h_{\varphi}(\sum_{i=1}^K g_{i} f_{\phi_i}( \cdot))$, where the optimizable MCL parameters $\theta_{\rm{MCL}}$ include $\bigcup_{i=1}^{K}\phi_i$, $\bigcup_{i=1}^{K}g_{i}$ and $\varphi$. 
The proposed MCL is applicable to a wide range of loss functions for continual learning. By default, here we focus on the synaptic regularization methods \cite{kirkpatrick2017overcoming,aljundi2018memory,zenke2017continual,chaudhry2018riemannian} as defined in Eq.~\eqref{eq:Regularization}. 
In order to coordinate the diversity of learners' expertise, we implement the proposed active forgetting in each learner. We further regularize differences in their predictive distributions, quantified by the widely-used Kullback Leibler (KL) divergence. Therefore, the loss function for the full version of our approach is defined as:

\begin{small}
\begin{equation}
\begin{split}
\mathcal{L}_{\rm{CAF}}(\theta_{\rm{MCL}}) & =  
\mathcal{L}_{\rm{t}}(\theta_{\rm{MCL}}) + \underbrace{\frac{\lambda_{\rm{SP}}}{2}\sum_{i=1}^{K}\sum_{m=1}^{M_i} \xi_{1:t-1,i,m} (\theta_{i,m} - \theta_{1:t-1,i,m}^*)^2}_{\rm{Stability \,\, Protection}} \\
& + \underbrace{ \sum_{i=1}^{K} \frac{\lambda_{{\rm{AF}},i}}{2} \sum_{m=1}^{M_i} (\theta_{i,m})^2}_{\rm{AF-1}} 
+ \underbrace{\sum_{i = 1, j \neq i}^{K} \frac{\gamma_{i,j}}{N_t} \sum_{n = 1}^{N_t} p_i(x_{t,n}) \log \frac{p_i(x_{t,n})}{p_j(x_{t,n})}}_{\rm{AF-S}},
\end{split}
\label{eq:CAF_loss} 
\end{equation}
\end{small}
\noindent
where $M_i$ denotes the amount of parameters $\theta_i = \{\phi_i, g_i\}$ for learner $i$. The current task has $N_t$ training samples, and $p_i(x_{t,n})$ is the prediction of learner $i$ for $x_{t,n}$.
We mainly consider AF-1 instead of AF-2 for computational efficiency, while keeping $\lambda_{\rm{SP}}$ identical to each learner for ease of implementation. 
We then discuss the implementation of $\lambda_{{\rm{AF}}, i}$ and $\gamma_{i,j}$. 
If we consider CAF as an overall continual learning model, active forgetting can be implemented similarly to Eq.~\eqref{eq:New_Posterior} and Eq.~\eqref{eq:AFEC_objective}, setting a uniform forgetting rate $\lambda_{{\rm{AF}}, i}$ as well as a uniform learning rule $\gamma_{i,j}$ to each learner.
This implementation is indeed effective under strong innate randomness to reduce excessive diversity .
In a more general context, the proposed MCL provides a spatial degree of freedom to modulate the diversity of learner's expertise. 
This idea can be implemented by constraining the average of $\{\lambda_{{\rm{AF}},i}\}_{i=1}^K$ to be a deterministic hyperparameter $\lambda_{\rm{AF}}$, i.e., $\frac{1}{K} \sum_{i=1}^{K} \lambda_{{\rm{AF}}, i}=\lambda_{\rm{AF}}$ where $\lambda_{{\rm{AF}}, i} = \alpha_i K \lambda_{\rm{AF}}$ and $\sum_{i=1}^{K} \alpha_i = 1$, but allowing their relative strength $\alpha_i$ as well as $\theta_{\rm{MCL}}$ to be optimized with gradients of the same loss function. Specifically, we perform softmax of a few optimizable parameters to ensure the constraint $\sum_{i=1}^{K} \alpha_i = 1$ and obtain $\alpha_i$ for each learning module. 
$\gamma_{i,j}$ can be implemented in a similar way by constraining their average $\frac{1}{K(K-1)} \sum_{i=1,j \neq i}^{K} \gamma_{i,j}=\gamma$, so as to enjoy the spatial degree of freedom.

\subsection{Theoretical Analysis of Generalization Ability\label{method_ta}}

Continual learning aims to find a solution $\theta$ that can generalize well over a new distribution $\mathbb{D}_t$ and a set of old distributions $\mathbb{D}_{1:t-1}:=\{\mathbb{D}_k\}_{k=1}^{t-1}$.
Let $\mathcal{E}_{\mathbb{D}_t}(\theta)$ and $\mathcal{E}_{\mathbb{D}_{1:t-1}}(\theta)$ denote the generalization errors. 
The training set and test set of each task follow the same distribution $\mathbb{D}_k $ ($k=1,2,...,t$), where the training set $D_{k} = \{(x_{k,n}, y_{k,n})\}_{n=1}^{{N}_{k}} $ includes $N_k$ data-label pairs. 
Then we define $\mathcal{E}_{\mathbb{D}_t}(\theta) = \mathbb{E}_{(x,y)\sim\mathbb{D}_t}[\mathcal{L}_t(\theta;x,y)]$ and $\mathcal{E}_{\mathbb{D}_{1:t-1}}(\theta) = \frac{1}{t-1} \sum_{k=1}^{t-1}  \mathbb{E}_{(x, y)\sim \mathbb{D}_k}[\mathcal{L}_k(\theta;x,y)]$, where $\mathcal{L}_k(\theta)$ can be generalized for any bounded loss function of task $k$. 
To minimize $\mathcal{E}_{\mathbb{D}_t}(\theta)$ and $\mathcal{E}_{\mathbb{D}_{1:t-1}}(\theta) $ without the use of old training samples $D_{1:t-1}:=\{D_k\}_{k=1}^{t-1}$, a continual learning model can only minimize an empirical risk over the current training samples ${D_t}$ in a parameter space $\Theta$ applicable to old tasks \cite{wang2022coscl,knoblauch2020optimal}, denoted as $\min_{\theta \in \Theta} \hat{\mathcal{E}}_{D_t} (\theta)$ where $\hat{\mathcal{E}}_{D_t} (\theta) = \frac{1}{N_t} \sum_{n=1}^{N_t} \mathcal{L}_t(\theta;x_{t,n},y_{t,n})$.
In practice, sequential learning of each task by $\min_{\theta \in \Theta} \hat{\mathcal{E}}_{D_t} (\theta)$ can find multiple solutions with different generalizability for $\mathcal{E}_{\mathbb{D}_t}(\theta)$ and $\mathcal{E}_{\mathbb{D}_{1:t-1}}(\theta)$, where a solution with flatter loss landscape typically acquires better generalizability and is therefore more robust to catastrophic forgetting \cite{deng2021flattening,cha2020cpr,mirzadeh2020understanding}. 

Accordingly, we define a robust empirical risk for the current task as $\hat{\mathcal{E}}_{D_t}^b (\theta) := \rm{max}_{\lVert \Delta \rVert \leq b} \hat{\mathcal{E}}_{\textit{D}_t} (\theta +\Delta)$ by the worst case of parameter perturbations $\Delta$, where $\lVert \cdot \rVert$ denotes the $L_2$ norm and $b$ is the radius of perturbations around $\theta$.
Likewise, a robust empirical risk for the old tasks is
$\hat{\mathcal{E}}_{D_{1:t-1}}^b (\theta) := \rm{max}_{\lVert \Delta \rVert \leq b} \hat{\mathcal{E}}_{\textit{D}_{1:t-1}} (\theta +\Delta)$. 
Then, $\min_{\theta \in \Theta} \hat{\mathcal{E}}_{{D}_t}^b (\theta)$ can find a flat minima over the current task. However, parameter changes that are much larger than the ``radius'' of the old minima can interfere with the performance of old tasks, while staying around the old minima can interfere with the performance of new tasks. Therefore, it is necessary to find a solution that can properly balance memory stability with learning plasticity, while being adequately compatible with the observed distributions. Formally, we analyze the generalization errors of a certain solution for continual learning with PAC-Bayes theory \cite{mcallester1999pac}, so as to provide the objective for computational modeling of neurological adaptive mechanisms. 
We leave technical details to Supplementary Section~1.3 and present the main results below: 

\vspace{-.2cm}
\begin{proposition}\label{CAF}
Let $\{\Theta_i \in  \mathbb{R}^{M_i}\}_{i=1}^{K}$ be a set of parameter spaces ($K\geq 1$ in general), 
$d_i$ be a VC dimension of $\Theta_i$, $\Theta = \cup_{i=1}^{K}\Theta_i$ with VC dimension $d$, and $M = \sum_{i=1}^{K} M_i$ as a given parameter budget.
Let $\hat \theta_{1:t}^{b}$ denote the optimal solution of the continually learned $1:t$ tasks by
robust empirical risk minimization over the current task, i.e., $\hat \theta_{1:t}^{b} = \arg \min_{\theta \in \Theta} \hat{\mathcal{E}}_{D_t}^b (\theta)$. Then for any $\delta \in (0,1)$, with probability at least $1-\delta$:

\begin{small}
\begin{align} \label{pro2-1}
\begin{split}
 \underbrace {\mathcal{E}_{\mathbb{D}_{t}} (\hat \theta_{1:t}^{b}) - \min_{\theta \in \Theta} \mathcal{E}_{\mathbb{D}_{t}} (\theta ) }_{\rm{Learning \,\, Plasticity}}
\leq  \underbrace {\min_{\theta \in \Theta} \hat{\mathcal{E}}_{D_{1:t-1}}^b (\theta) -  \min_{\theta \in \Theta} \hat{\mathcal{E}}_{D_{1:t-1}} (\theta)}_{\rm{Loss \,\, Flatness}} + 
\underbrace {\frac{1}{t-1}\sum_{k=1}^{t-1} \rm{Div}( \mathbb{D}_\textit{k}, \mathbb{D}_\textit{t})}_{\rm{Task \,\, Discrepancy}} + C_1, \\
 \underbrace {\mathcal{E}_{\mathbb{D}_{1:t-1}} (\hat \theta_{1:t}^{b}) - \min_{\theta \in \Theta} \mathcal{E}_{\mathbb{D}_{1:t-1}} (\theta ) }_{\rm{Memory \,\, Stability}}
\leq \underbrace {\min_{\theta \in \Theta} \hat{\mathcal{E}}_{D_t}^b (\theta) -  \min_{\theta \in \Theta} \hat{\mathcal{E}}_{D_t} (\theta)}_{\rm{Loss \,\, Flatness}} +  \underbrace {\frac{1}{t-1}\sum_{k=1}^{t-1} \rm{Div}(\mathbb{D}_\textit{t}, \mathbb{D}_\textit{k})}_{\rm{Task \,\, Discrepancy}} +C_2,
 \end{split}
\end{align}
\end{small}

\noindent
where 
$ C_{1} = \max_{i \in [1,K]} \sqrt{\frac{d_i\ln (N_{1:t-1}/d_i) + \ln (2K/\delta)}{N_{1:t-1}} }+ \sqrt{\frac{d\ln (N_{1:t-1}/d) + \ln (2/\delta)}{N_{1:t-1}} }$ and
$ C_{2} = \max_{i \in [1,K]} \sqrt{\frac{d_i\ln (N_{t}/d_i) + \ln (2K/\delta)}{N_{t}} } + \sqrt{\frac{d\ln (N_{t}/d) + \ln (2/\delta)}{N_{t}} }$ represent the cover of parameter space.
$\rm{Div}(\mathbb{D}_\textit{i}, \mathbb{D}_\textit{j}) := 2 \sup_{\textit{h} \in \mathcal{H}} \vert\mathcal{P}_{ \mathbb{D}_\textit{i}}(\textit{I}(\textit{h})) - \mathcal{P}_{ \mathbb{D}_\textit{j}}\textit{I}(\textit{h}))\vert$ is the $\mathcal{H}$-divergence of $\mathbb{D}_i$ and $\mathbb{D}_j$, where $I(h)$ is the characteristic function. 
$N_{1:t-1}=\sum_{k=1}^{t-1} N_k$ is the total number of training samples over all old tasks, where $N_k$ is the number of training samples over task $k$.
\end{proposition}
\vspace{-.3cm}

From Proposition~\ref{CAF}, the generalization gaps over new and old tasks, corresponding to learning plasticity and memory stability, are uniformly constrained by the loss flatness and task discrepancy, which further depend on the cover of parameter space, i.e., $ C_{1}$ and $ C_{2}$.
In particular, we have $C_1 = 2\sqrt{\frac{d\ln (N_{1:t-1}/d) + \ln (2/\delta)}{N_{1:t-1}} }$ and $C_2 = 2\sqrt{\frac{d\ln (N_{t}/d) + \ln (2/\delta)}{N_{t}} }$ for $K=1$. 
When $K>1$, $C_1 < 2\sqrt{\frac{d\ln (N_{1:t-1}/d) + \ln (2/\delta)}{N_{1:t-1}} }$ and $C_2 < 2\sqrt{\frac{d\ln (N_{t}/d) + \ln (2/\delta)}{N_{t}} }$ due to $d_i < d$ for $i \in [1,K]$.
This means that employing multiple continual learners (i.e., $K>1$) compared to a single continual learner (i.e., $K=1$) can tighten the generalization bounds and thus benefit the performance of continual learning. 
Likewise, the benefits of active forgetting can be explained from a similar perspective:

\begin{proposition} \label{AF}
Let $\{\Theta_i \in  \mathbb{R}^{M_i}\}_{i=1}^{K}$ be a set of parameter spaces ($K\geq 1$ in general), 
$d_i$ be a VC dimension of $\Theta_i$, $\Theta = \cup_{i=1}^{K}\Theta_i$ with VC dimension $d$, and $M = \sum_{i=1}^{K} M_i$ as a given parameter budget.
Based on Proposition~\ref{CAF}, for $\hat \theta_{1:t}^{b} = \arg \min_{\theta \in \Theta} \hat{\mathcal{E}}_{D_t}^b (\theta)$, the upper bound of generalization gap is further externalized with  
$C_{1} = r_1(\lVert \hat \theta_{1:t}^{b} \rVert_{2}^2/b^2)$
and
$C_{2} = r_2(\lVert \hat \theta_{1:t}^{b} \rVert_{2}^2/b^2)$,
where $r_1: \mathbb{R}_+ \rightarrow \mathbb{R}_+$ and $r_2: \mathbb{R}_+ \rightarrow \mathbb{R}_+$ are two strictly increasing functions under some technical conditions on $\mathcal{E}_{\mathbb{D}_{t}}(\theta) $ and $\mathcal{E}_{\mathbb{D}_{1:t-1}}(\theta) $, respectively.
\end{proposition}  
\vspace{-.3cm}

Proposition~\ref{AF} externalizes the two generalization bounds in Proposition~\ref{CAF} by defining the VC dimension with $L_2$ norm for parameters under some technical assumptions.
Notably, the claim in Proposition~\ref{CAF} regarding the benefits of using multiple continual learning still holds in such a tightened version. 
In particular, optimization of the active forgetting term in Eq.~\eqref{eq:AFEC_objective} and Eq.~\eqref{eq:CAF_loss}, which takes the form of minimizing a weighted $L_2$ norm regarding all parameters, can contribute to tightening the two generalization bounds in Proposition~\ref{AF}, especially through optimizing $C_1$ and $C_2$.
Besides, $N_{1:t-1}$ becomes larger as $t$ increases, while $N_{t}$ remains constant in general.
This means that $C_1$ will decrease more rapidly around the empirical optimal solution $\hat \theta_{1:t}^{b} = \arg \min_{\theta \in \Theta} \hat{\mathcal{E}}_{D_t}^b (\theta)$ as more tasks are introduced, resulting in more pronounced improvements in learning plasticity.

\subsection{Implementation\label{method_imp}}

We mainly consider the setting of task-incremental learning \cite{van2022three} to perform continual learning experiments, with task identities provided in both training and testing.
We split representative benchmark datasets for visual classification tasks. The CIFAR-100 dataset \citep{krizhevsky2009learning} includes 100-class colored images of size $32 \times 32$. We split it based on different principles to evaluate the effect of overall knowledge transfer. Specifically, R-CIFAR-100 and S-CIFAR-100 are constructed by splitting CIFAR-100 into 20 tasks, depending on random order or superclasses defined by semantic similarity, respectively. R-CIFAR-10/100 includes 2 tasks randomly split from the CIFAR-10 dataset \citep{krizhevsky2009learning} of 10-class colored images, followed by the 20 tasks of R-CIFAR-100. The Omniglot dataset \cite{lake2015human} includes 50 alphabets for a total of 1623 classes of characters, where each class contains 20 hand-written digits of size \(28 \times 28\). We split each alphabet as a task consisting of a different number of classes. 
The CUB-200-2011 dataset \cite{wah2011caltech} includes 200-class bird images of size $224 \times 224$, and the Tiny-ImageNet dataset \cite{delange2021continual} includes 200-class natural images of size \(64 \times 64\), both split randomly into 10 tasks. The CORe50 dataset \cite{lomonaco2017core50} includes 50 handheld objects with smoothly-changed observations of size $128 \times 128$, randomly split into 10 tasks \cite{pham2021contextual}.
We construct a sequence of Atari reinforcement tasks for continual learning, i.e., DemonAttack - Robotank - Boxing - NameThisGame - StarGunner - Gopher - VideoPinball - Crazyclimber, using the same PPO algorithm \cite{schulman2017proximal} to learn each task. The network architecture and training regime are described in Supplementary Sections~2.1 and~2.2, respectively.

\subsection{Baseline Approach}\label{sec4.5}
To ensure generality in realistic applications, we restrict the old training samples to be unavailable in continual learning, and compare with representative methods that follow this restriction. Specifically, EWC \cite{kirkpatrick2017overcoming}, MAS \cite{aljundi2018memory} and SI \cite{zenke2017continual} are synaptic regularization methods that selectively penalized parameter changes to preserve memory stability. AGS-CL \cite{jung2020continual} took advantages of parameter isolation and synaptic regularization to prevent catastrophic forgetting. P\&C \cite{schwarz2018progress} adopted an additional active column on the basis of EWC \cite{kirkpatrick2017overcoming} to improve learning plasticity. CPR \cite{cha2020cpr} encouraged convergence to a flat loss landscape, which can be combined with other baseline approaches. The hyperparameters for continual learning are determined with a comprehensive grid search. We construct a different task sequence (e.g., different class splits, data shuffling, task orders and random seeds) from the actual experiments and run it for once. 
Then we use the best combinations of hyperparameters to perform the actual experiments for multiple runs, as described in Supplementary Section~2.3.

\subsection{Evaluation Metric}
We consider three evaluation metrics for visual classification tasks, i.e., average accuracy (AAC), forward transfer (FWT) and backward transfer (BWT) \cite{lopez2017gradient,wang2023comprehensive}:
\begin{small}
\begin{equation}
    {\rm{AAC}} = \frac{1}{T} \sum_{i=1}^{T} {\rm{A}}_{T,i},
\end{equation}
\begin{equation}
    {\rm{FWT}} = \frac{1}{T-1} \sum_{i=2}^{T} {\rm{A}}_{i-1,i} - \tilde{a}_{i},
\end{equation}
\begin{equation}
    {\rm{BWT}} = \frac{1}{T-1} \sum_{i=1}^{T-1} {\rm{A}}_{T,i} - {\rm{A}}_{i,i},
\end{equation}
\end{small}
where \({\rm{A}}_{t,i}\) is the test accuracy of task \(i\) after continual learning of task \(t\), and \(\tilde{a}_{i}\) is the test accuracy of each task \(i\) learned from random initialization. ACC is the average performance of all tasks ever seen, which evaluates the overall performance of continual learning. FWT evaluates the average influence of remembering old tasks to new tasks for learning plasticity. BWT evaluates the average influence of learning new tasks to old tasks for memory stability. 

The diversity of learners' predictions is quantified by the average cosine (Cos) or Euclidean (Euc) distance: 
\begin{small}
\begin{equation}
{\rm{Cos}} = 1 - \frac{1}{K(K-1)} \sum_{i=1,j \neq i}^{K} \frac{p_i \cdot p_j}{\Vert p_i \Vert \cdot \Vert p_j \Vert}, 
\end{equation}
\begin{equation}
{\rm{Euc}} = \frac{1}{K(K-1)} \sum_{i=1,j \neq i}^{K} \Vert p_i - p_j \Vert,
\end{equation}
\end{small}
where $p_i$ and $p_j$ denote the predictions of learners $i$ and $j$, respectively.

The discrepancy of task distributions in feature space is evaluated by the difficulty of distinguishing them, which is an empirical approximation of the $\mathcal{H}$-divergence \cite{long2015learning,wang2022coscl} in Proposition~\ref{CAF}.
We train a simple discriminator consisting of a fully-connected layer and use binary cross-entropy to measure the average discrimination error on test sets. 

The performance of Atari reinforcement tasks is evaluated by the normalized accumulated reward (NAR), normalized plasticity (NP) and normalized stability (NS), corresponding to the overall performance, learning plasticity and memory stability, respectively:

\begin{small}
\begin{equation}
    {\rm{NAR}} = \sum_{i=1}^{T} {\rm{R}}_{T,i}/{r}_{i},
\end{equation}
\begin{equation}
    {\rm{NP}} = \frac{1}{T-1} \sum_{i=2}^{T} {\rm{R}}_{i,i}/{r}_{i},
\end{equation}
\begin{equation}
    {\rm{NS}} = \frac{1}{T-1} \sum_{i=1}^{T-1} {\rm{R}}_{T,i} / {\rm{R}}_{i,i},
\end{equation}
\end{small}
where ${\rm{R}}_{t,i}$ is the reward for task $i$ obtained in the test step after learning task $t$, and $r_{i}$ is the maximum reward for task $i$ obtained in each test step of fine-tuning on the task sequence.

\backmatter

\section*{Data Availability}
All benchmark datasets used in this paper are publicly available, including CIFAR-10/100 \citep{krizhevsky2009learning} (https://www.cs.toronto.edu/~kriz/cifar.html), Omniglot \cite{lake2015human} (https://www.omniglot.com), CUB-200-2011 \cite{wah2011caltech} (https://www.vision.caltech.edu/datasets/cub\_200\_2011/), Tiny-ImageNet \cite{delange2021continual} (https://www.image-net.org/download.php), CORe50 \cite{lomonaco2017core50} (https://vlomonaco.github.io/core50/) and Atari games \cite{mnih2013playing} (https://github.com/openai/baselines).

\section*{Code Availability}
The implementation code is available at the GitHub repository https://github.com/lywang3081/CAF\cite{wang2023caf_code}.

\section*{Acknowledgments}
This work was supported by the National Key Research and Development Program of China (2020AAA0106302, to J.Z.), the STI2030-Major Projects (2022ZD0204900, to Y.Z.), the National Natural Science Foundation of China (Nos. 62061136001 and 92248303, to J.Z., 32021002, to Y.Z., U19A2081, to H.S.), the Tsinghua-Peking Center for Life Sciences, the Tsinghua Institute for Guo Qiang, and the High Performance Computing Center, Tsinghua University. J.Z. was also supported by the New Cornerstone Science Foundation through the XPLORER PRIZE. L.W. was also supported by Shuimu Tsinghua Scholar.

\section*{Author Contributions Statement}
L.W., X.Z., J.Z. and Y.Z. conceived the project. L.W., X.Z., Q.L. and M.Z. designed the computational model. X.Z. performed the theoretical analysis, assisted by L.W. L.W. performed all experiments and analyzed the data. L.W., X.Z. and Q.L. wrote the paper. L.W., X.Z., Q.L., M.Z., H.S., J.Z. and Y.Z. revised the paper. J.Z. and Y.Z. supervised the project. 

\section*{Competing Interests Statement}
The authors declare no competing interests.

\clearpage
\begin{figure}[t]
    \centering
    \includegraphics[width=1\linewidth]{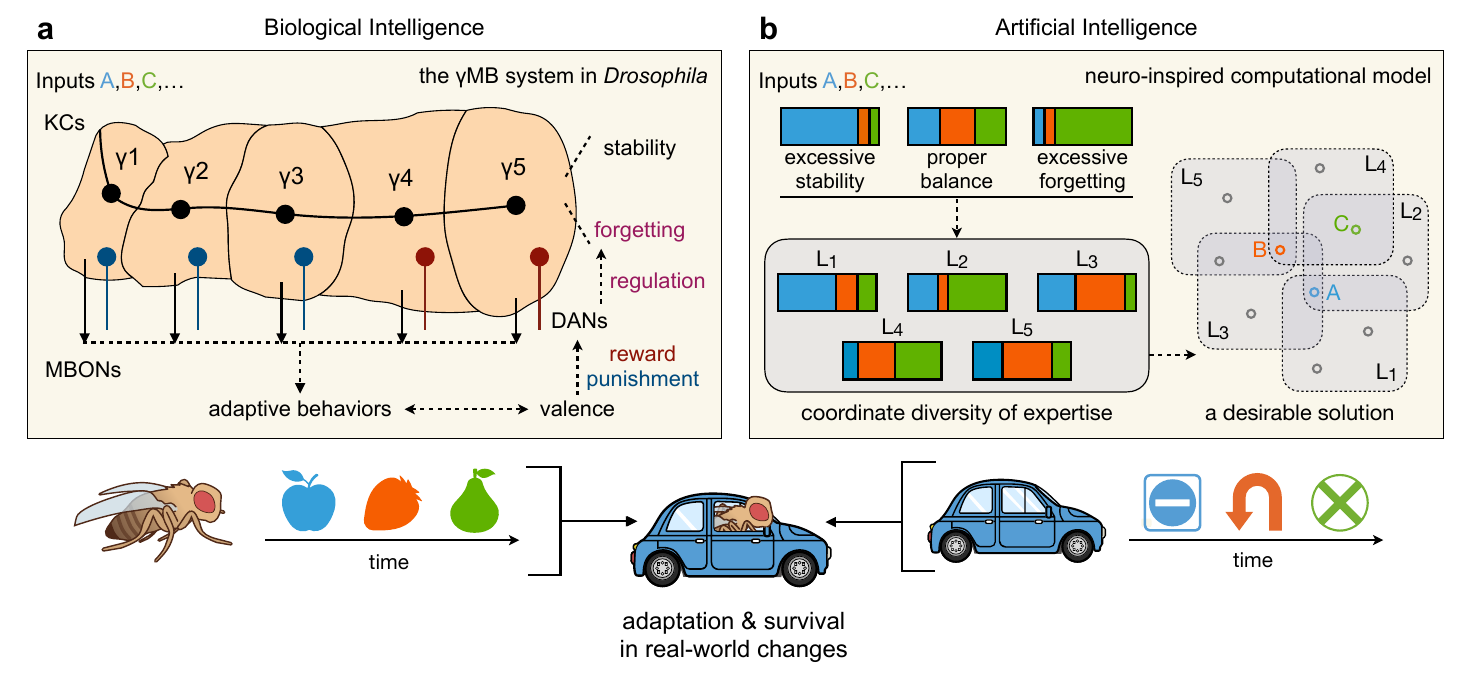}
\caption{\textbf{Continual learning with reference to a biological learning system.} \textbf{a}, The \textit{Drosophila} $\gamma$MB system has evolved adaptive mechanisms to cope with different tasks in succession, such as selective stabilization of synaptic changes, active regulation of memory decay (i.e., active forgetting) and dynamic coordination of multiple parallel compartments (i.e., $\gamma$1-5).
KCs: Kenyon cells; DANs: dopaminergic neurons; MBONs: mushroom body output neurons. \textbf{b}, Inspired by such biological strategies, we propose to incorporate active forgetting together with stability protection for a better trade-off between new and old tasks, and accordingly coordinate multiple continual learners to ensure solution compatibility.
${\rm{L}}_{1-5}$: five continual learners corresponding to the five compartments. The dashed areas on the lower right denote the target distributions of ${\rm{L}}_{1-5}$, and the small hollow circles represent the optimal solution for each incremental task (tasks $A$, $B$, $C$ are colored, and other tasks are gray).} 
\label{Figure1}
\end{figure}

\begin{figure}[t]
    \centering
    \vspace{-0.1cm}
    \includegraphics[width=1\linewidth]{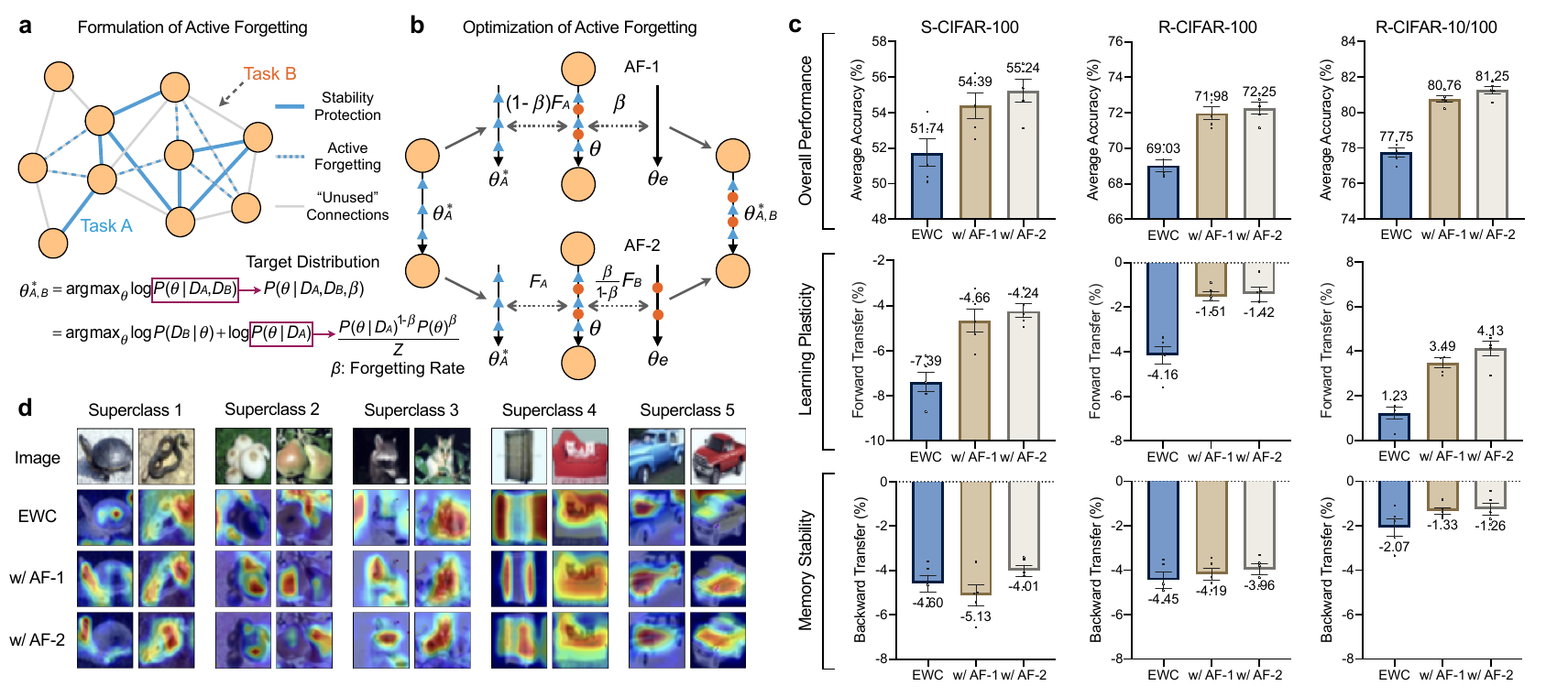}
\caption{\textbf{Implementation of active forgetting in a continual learning model.} \textbf{a}, Formulating active forgetting together with stability protection through the framework of Bayesian learning. \textbf{b}, The proposed functional strategy can be optimized in two equivalent ways of synaptic expansion-renormalization (AF-1 and AF-2), where the network parameters $\theta$ need to be selectively renormalized with both $\theta_A^*$ and $\theta_e$ in order to balance new and old tasks mutually in a shared solution.
\textbf{c}, Experimental results. The evaluation metrics include average accuracy for overall performance (top), forward transfer for learning plasticity of new tasks (middle) and backward transfer for memory stability of old tasks (bottom). Because of different construction principles, the overall knowledge transfer ranges from more negative to more positive across S-CIFAR-100, R-CIFAR-100 and R-CIFAR-10/100. We use a 6-layer convolution neural network (CNN) similar to the previous work \cite{jung2020continual,wang2021afec,cha2020cpr,wang2022coscl}. All results are averaged over 5 runs with different random seeds and task orders. The error bars represent the standard error of the mean. \textbf{d}, Visualization of the latest task predictions on S-CIFAR-100 with Grad-CAM \cite{selvaraju2017grad}.}
\label{Figure2}
\end{figure}

\begin{figure}[t]
    \centering
    \vspace{-0.1cm}
    \includegraphics[width=1\linewidth]{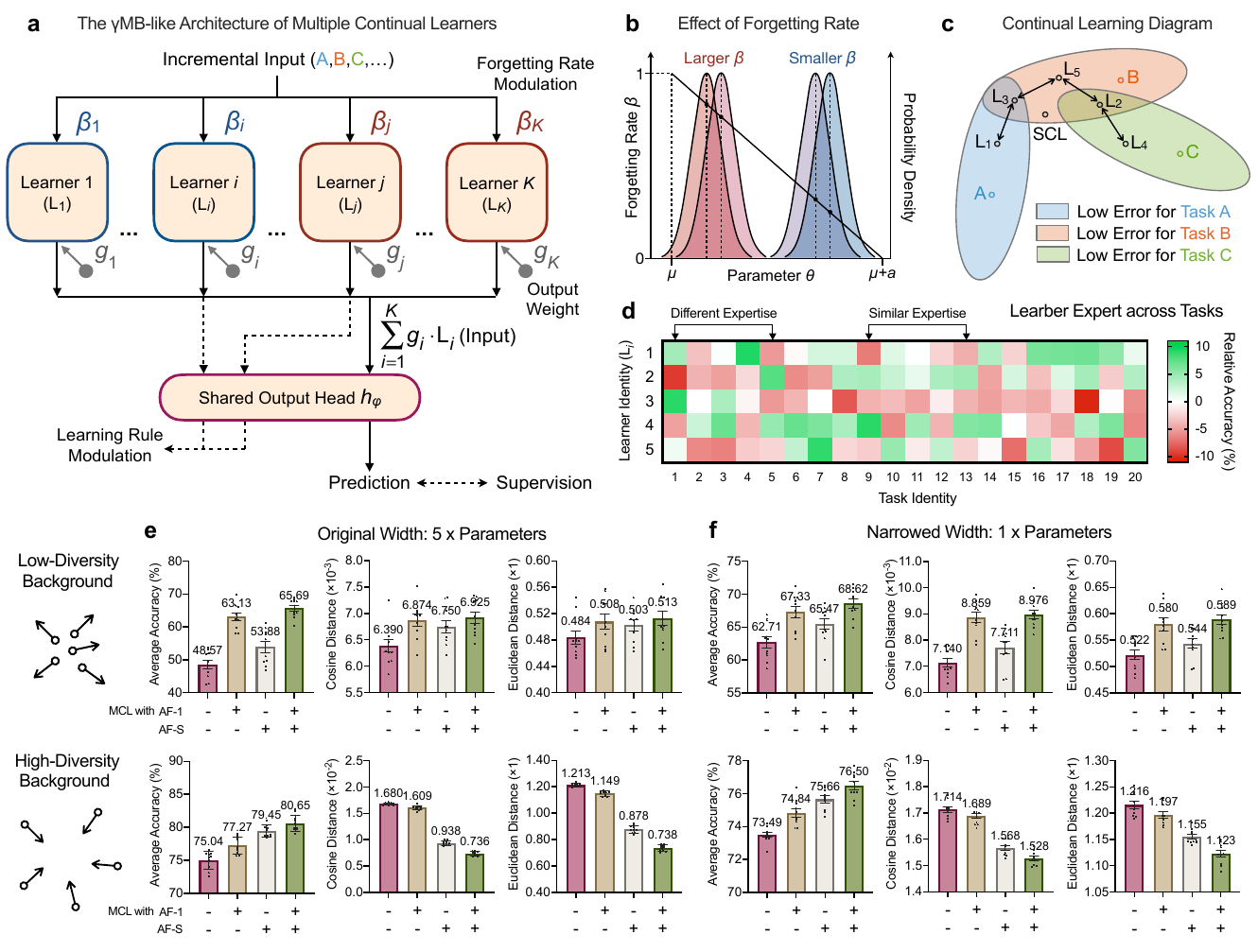}
\caption{\textbf{The $\gamma$MB-like architecture with adaptive modulations.} \textbf{a}, Inspired by the organizing principles of the $\gamma$MB system, we design a specialized architecture consisting of multiple parallel continual learners ${\rm{L}}_{1-K}$. \textbf{b}, Learner differences can be modulated by the forgetting rate $\beta$. Here we present an example for the target distribution $p(\theta \vert D_{A}, D_{B}, \beta)$ when $p(\theta) \sim \mathcal{N}(\mu, \sigma^2)$ and $p(\theta \vert D_{A}) \sim \mathcal{N}(\mu+a, \sigma^2)$ in $\hat{p}(\theta \vert D_A, \beta) = \frac{p(\theta \vert D_A)^{(1 - \beta) } p(\theta)^{\beta }}{Z}$, where $p(\theta \vert D_{A}, D_{B}, \beta)$ is also a Gaussian (Supplementary~\ref{secA_af_posterior}) and the vertical dashed line denotes its mode. 
\textbf{c}, A conceptual diagram of continual learning. $A$, $B$, and $C$ are three incremental tasks with different similarities. ${\rm{L}}_{1-5}$ are five learners with proper diversity in parameter space.
The colored hollow circles indicate the optimal solution for each task. 
\textbf{d}, Learners' expertise across tasks. After continual learning of all tasks, we evaluate the relative accuracy of each learner across tasks, calculated as the performance of each learner minus the average performance of all learners. 
\textbf{e}, \textbf{f}, The two adaptive modulations (i.e., AF-1 and AF-S) can improve the performance of MCL to a large extend through coordinating the diversity of learners' expertise, as measured by the average cosine or Euclidean distance of their predictions. All experiments are performed on R-CIFAR-100 with EWC \cite{kirkpatrick2017overcoming} as the baseline approach. The results in \textbf{e}, \textbf{f} are averaged over 10 runs with different random seeds and task orders. The error bars represent the standard error of the mean.}
\label{Figure3}
\end{figure}

\begin{figure}[t]
    \centering
    \vspace{-0.1cm}
    \includegraphics[width=0.90\linewidth]{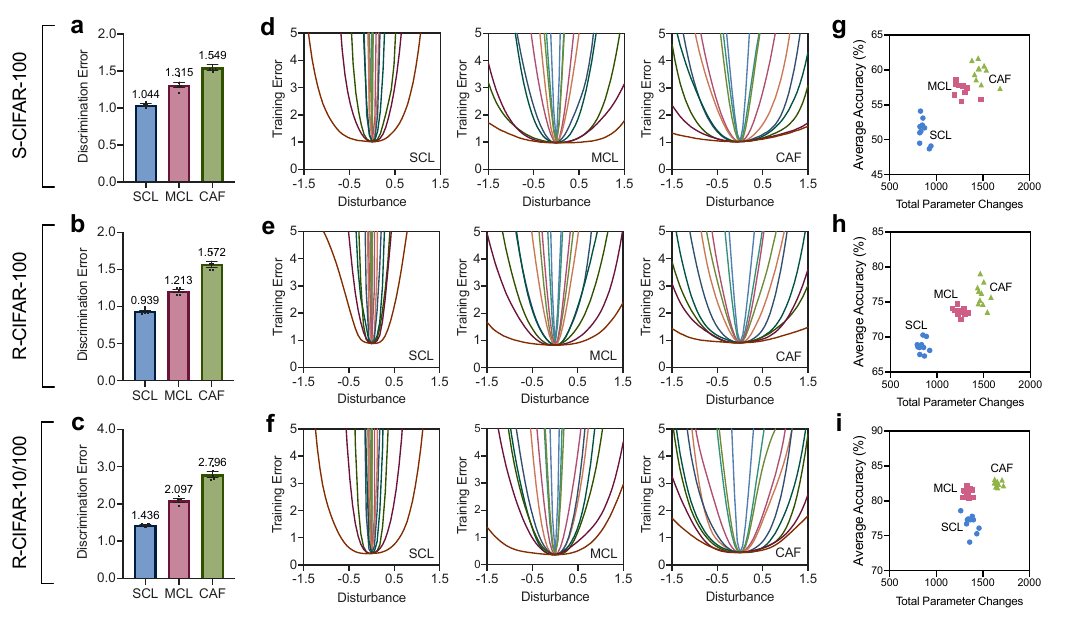}
\caption{\textbf{Exploration of underlying mechanisms that improve continual learning.}
We build the MCL with a narrowed width to keep the total amount of parameters similar to that of the SCL, and adopt the high-diversity background as a sub-optimal implementation. All experiments are performed with EWC \cite{kirkpatrick2017overcoming} as the baseline approach.
\textbf{a}-\textbf{c}, We empirically approximate the discrepancy of task distributions in feature space by training a simple discriminator to distinguish whether the feature of an input image belongs to a task or not, and measure the discrimination error on test set via binary cross-entropy, where a larger discrimination error indicates a smaller difference \cite{long2015learning,wang2022coscl}. The error bars represent the standard error of the mean over 5 runs.
\textbf{d}-\textbf{f}, Curvature of loss landscape around the obtained solution after continual learning of all tasks. After disturbing the network parameters with random values, we evaluate the training error with cross-entropy \cite{deng2021flattening}, where each line is a different direction of disturbance. 
\textbf{g}-\textbf{i}, Total parameter changes and average accuracy of all tasks. Each point represents a run.
} 
\label{Figure4}
\end{figure}

\begin{figure}[t]
    \centering
    \vspace{-0.1cm}
    \includegraphics[width=0.95\linewidth]{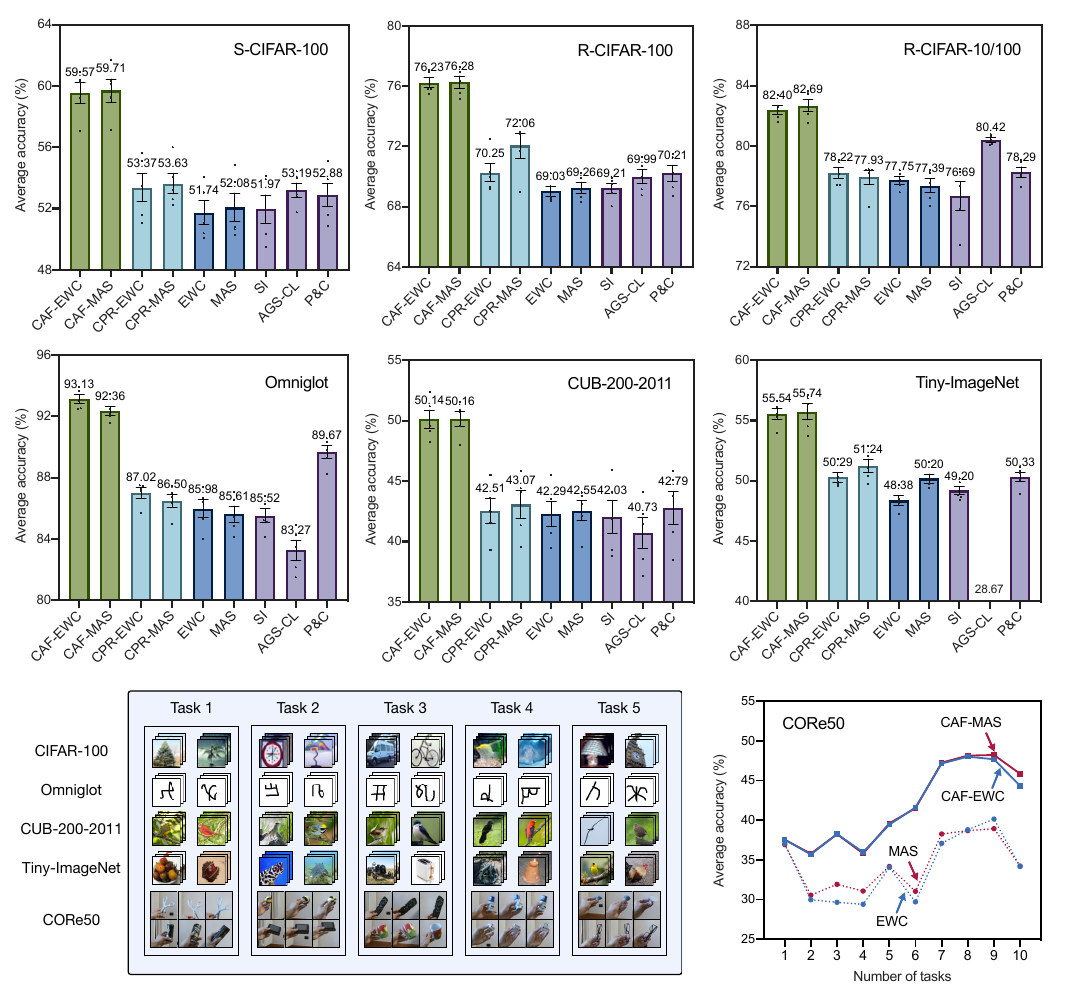}
\caption{\textbf{Performance evaluation for visual classification tasks.} 
We consider seven visual classification benchmarks to evaluate different aspects of continual learning, such as overall knowledge transfer, input size, number of training samples, length of task sequence, smoothly-changed observations, etc. Please refer to the sub-figure below left for a demo of incremental tasks.
CAF (ours) and CPR \cite{cha2020cpr} are plug-and-play for regularization-based methods such as EWC \cite{kirkpatrick2017overcoming} and MAS \cite{aljundi2018memory}. 
Under similar parameter budgets, all results are averaged over 5 runs with different random seeds and task orders. The error bars represent the standard error of the mean.
} 
\label{Figure5}
\end{figure}

\begin{figure}[t]
    \centering
    \vspace{-0.1cm}
    \includegraphics[width=0.70\linewidth]{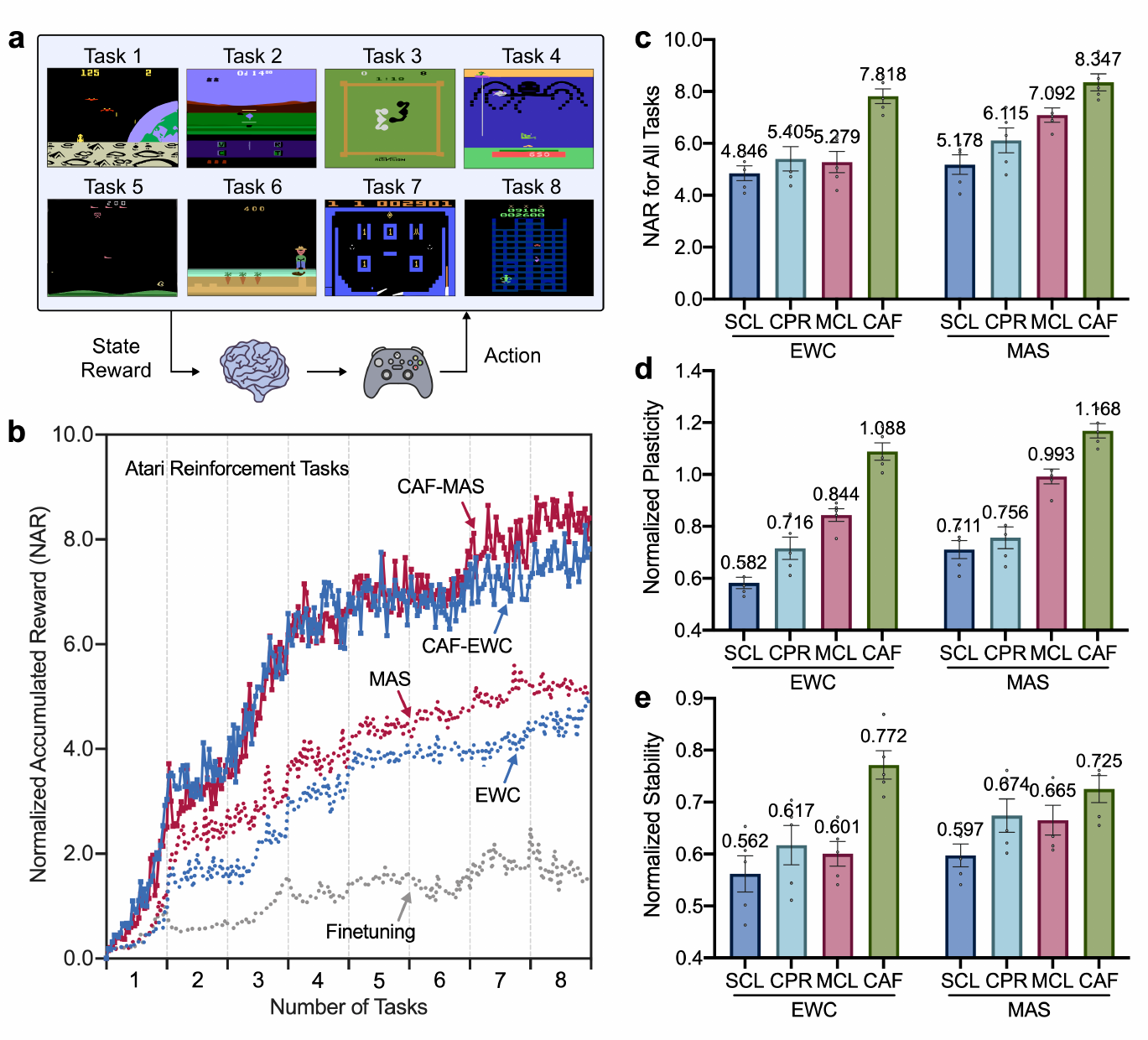}
    \caption{\textbf{Performance evaluation for Atari reinforcement tasks.} 
    \textbf{a}, An intelligent agent attempts to acquire more rewards from learning a sequence of Atari games.
    \textbf{b}, The normalized accumulated reward (NAR) in continual learning. The performance of simply fine-tuning on the task sequence is used to normalize the reward obtained for each task.
    \textbf{c}-\textbf{e}, After continual learning of all Atari games, we evaluate the overall performance, learning plasticity and memory stability, respectively. 
    Under similar parameter budgets, all results are averaged over 5 runs with different random seeds. The error bars represent the standard error of the mean.
    } 
    \label{4}
    \vspace{-0.1cm}
\end{figure}

\clearpage


\bibliography{sn-bibliography}

\clearpage




\section*{Supplementary material}

# 1 Theoretical Foundation and Analysis

## 1.1 Posterior Distribution with Active Forgetting

For arbitrary Gaussian distributions $p_1(x) \sim \mathcal{N}(\mu_1, \sigma_1^2)$, $p_2(x) \sim \mathcal{N}(\mu_2, \sigma_2^2)$ and $\beta \in [0, 1]$, there exists a $\beta$-dependent normalizer $Z$ that keeps $\hat{p}(x) = \frac{p_1(x)^{1-\beta} p_2(x)^\beta}{Z}$ following a Gaussian distribution. Specifically, the probability density functions of $p_1(x)$ and $p_2(x)$ can be defined as

$$p_1(x) = \frac{1}{\sqrt{2\pi\sigma_1^2}} \exp(-\frac{(x-\mu_1)^2}{2\sigma_1^2}), \tag{S1}$$

$$p_2(x) = \frac{1}{\sqrt{2\pi\sigma_2^2}} \exp(-\frac{(x-\mu_2)^2}{2\sigma_2^2}). \tag{S2}$$

We get

$$
\begin{aligned}
p_1(x)^{1-\beta} p_2(x)^\beta &= \frac{1}{\sqrt{2\pi\sigma_1^{2(1-\beta)}\sigma_2^{2\beta}}} \exp(-[\frac{(1-\beta)(x-\mu_1)^2}{2\sigma_1^2} + \frac{\beta(x-\mu_2)^2}{2\sigma_2^2}]) \\
&= \frac{1}{\sqrt{2\pi\sigma_1^{2(1-\beta)}\sigma_2^{2\beta}}} \exp(-[\frac{(x-m)^2}{2v^2} + \frac{k-m^2}{2v^2}]),
\end{aligned}
\tag{S3}
$$

where

$$v^2 = \frac{\sigma_1^2 \sigma_2^2}{\beta\sigma_1^2 + (1-\beta)\sigma_2^2}, \tag{S4}$$

$$m = \frac{(1-\beta)\sigma_2^2\mu_1 + \beta\sigma_1^2\mu_2}{\beta\sigma_1^2 + (1-\beta)\sigma_2^2}, \tag{S5}$$

$$k = \frac{(1-\beta)\sigma_2^2\mu_1^2 + \beta\sigma_1^2\mu_2^2}{\beta\sigma_1^2 + (1-\beta)\sigma_2^2}. \tag{S6}$$

Then we get

$$\hat{p}(x) = \frac{p_1(x)^{1-\beta} p_2(x)^\beta}{Z} \sim \mathcal{N}(m, v^2), \tag{S7}$$

$$Z = \sqrt{\frac{v^2}{\sigma_1^{2(1-\beta)}\sigma_2^{2\beta}}} \exp(-\frac{k-m^2}{2v^2}). \tag{S8}$$

## 1.2 Implementation of Active Forgetting

Here we discuss in more depth the motivation and implementation of active forgetting, serving as an extension of its Methods. First, we would clarify that the use of "active forgetting" is borrowed from the corresponding biological mechanism [16, 6, 8, 13, 12, 5, 21], suggesting that the old memories are actively removed from network parameters. Similar to the biological function of active forgetting in providing flexibility, this proposal could enhance the probability of learning each new task well, as demonstrated in Methods Eq. (5). In practice, Eq. (5) is not actually used in experiments but rather to provide theoretical foundations for determining of an appropriate $\beta$. Because $\beta^*$ that maximizes the current task performance

may affect previous and subsequent tasks (e.g., excessive elimination of old knowledge), the selected forgetting rate should take into account the overall effect of continual learning, which is generally smaller than $\beta^*$ but still improves learning plasticity. In other words, Eq. (5) demonstrates that there generally exists an appropriate $\beta \in (0, \beta^*]$ to enhance learning plasticity in continual learning. We can then perform a grid search to determine it over the task sequence in experiments.

After incorporating active forgetting into continual learning, we derive two equivalent versions from Methods Eq. (6), i.e., AF-1 and AF-2 as described in Methods Eq. (9). The derivation of Eq. (9) is detailed as follows:

$$
\begin{aligned}
\theta_{A,B}^* &= \arg\max_\theta \log p(\theta | D_A, D_B, \beta) \\
&= \arg\max_\theta \log p(D_B | \theta) + \log \hat{p}(\theta | D_A, \beta) - \underbrace{\log p(D_B)}_{const.} \\
&= \arg\max_\theta \log p(D_B | \theta) + (1 - \beta) \log p(\theta | D_A) + \beta \log p(\theta) - \underbrace{(\log Z + \log p(D_B))}_{const.} \\
&\overset{\text{AF-1}}{=} \arg\max_\theta \log p(D_B | \theta) + (1 - \beta) \log p(\theta | D_A) + \beta \log p(\theta) \\
&= \arg\max_\theta (1 - \beta + \beta) \log p(D_B | \theta) + (1 - \beta) \log p(\theta | D_A) + \beta \log p(\theta) - \underbrace{\beta \log p(D_B)}_{const.} \\
&= \arg\max_\theta (1 - \beta) \log p(D_B | \theta) + (1 - \beta) \log p(\theta | D_A) + \beta[\log p(\theta) + \log p(D_B | \theta) - \log p(D_B)] \\
&\overset{\text{AF-2}}{=} \arg\max_\theta (1 - \beta) \log p(D_B | \theta) + (1 - \beta) \log p(\theta | D_A) + \beta \log p(\theta | D_B).
\end{aligned}
\tag{S9}
$$

For AF-1, $\theta_{e,m} = 0$, $I_{e,m} = 1$, $\lambda_{\text{AF}} \propto \beta$ and $\lambda_{\text{SP}} \propto (1 - \beta)$. For AF-2, $\theta_{e,m} = \theta_{t,m}^*$, $I_{e,m} = F_{t,m}$, $\lambda_{\text{AF}} \propto \beta/(1-\beta)$ and $\lambda_{\text{SP}} \propto 1$. AF-2 has an intuitive explanation that the network parameters $\theta$ need to be selectively renormalized with both new and old task solutions to achieve an appropriate trade-off between them, but results in more than twice the resource overhead to compute $\theta_e$ and $I_e$. In contrast, AF-1 is an implicit and efficient way of improving learning plasticity beyond preserving memory stability. For example, when using a larger $\beta$, AF-2 corresponds to increasing $\lambda_{\text{AF}}$ while keeping $\lambda_{\text{SP}}$ constant, in order to explicitly improve the new task performance. AF-1 corresponds to increasing $\lambda_{\text{AF}}$ while decreasing $\lambda_{\text{SP}}$, so as to implicitly improve the new task performance.

Interestingly, AF-1 and $L_2$ weight decay are different in motivation while similar in practical implementation. Specifically, the loss function of AF-1 adds a weighted $L_2$ penalty to the regular synaptic regularization methods as described in Methods Eq. (11), but with hyperparameters related to the forgetting rate $\beta$ as an implicit trade-off between new and old tasks. When $\lambda_{\text{SP}}$ and $\lambda_{\text{AF}}$ are independently varied, i.e., $\lambda_{\text{SP}}$ re-uses the optimal value of Eq. (11) and $\lambda_{\text{AF}}$ is determined by a new grid search, AF-1 becomes the same as $L_2$ weight

decay in practical implementation, with tolerable performance degradation compared to a grid search of $\lambda_{\text{SP}}$ and $\lambda_{\text{AF}}$ together (Supplementary Fig. 2). This provides an alternative way of implementing AF-1 to avoid compromising the comparison baselines. On the other hand, the commonly-used protocol for selecting hyperparameters from a grid search requires access to the full length of a task sequence, which can be problematic for continual learning in practice [4, 17] and may attribute the demonstrated improvements of AF-1 and AF-2 to the use of additional hyperparameters. To validate the effectiveness of active forgetting, we first perform a sensitivity-analysis of $\lambda_{\text{SP}}$ and $\lambda_{\text{AF}}$ in Supplementary Fig. 3a. As can be seen, both AF-1 and AF-2 largely improve the performance of EWC over a wide range of $\lambda_{\text{AF}}$. In particular, the $\lambda_{\text{AF}}$ that achieves comparably strong performance covers three orders of magnitude in all cases. Therefore, the performance improvements are remarkably robust to the changes of $\lambda_{\text{AF}}$. Meanwhile, we follow the idea of cross-validation [4] and analyze the impact of hyperparameters with a short sequence of 5 randomly-selected tasks. As shown in Supplementary Fig. 3, the "good regions" of cross-validation search are basically contained in the counterpart of regular grid search, although their performance patterns differ to some extent. In actual experiments, we have further verified that the performance of EWC w/ AF-1 is indeed comparable between regular grid search and cross-validation search, thanks to the robustness of hyperparameters.

## 1.3 Generalization Bound for Continual Learning

**Proof of Proposition 1:** To complete the proof of Proposition 1, we first present another two important propositions (i.e., Proposition 3 and Proposition 4) as follows.

**Proposition 3.** *Let $\Theta$ be a cover of a parameter space with VC dimension $d$. If $\mathbb{D}_1, \cdots, \mathbb{D}_t$ are the distributions of the continually learned $1 : t$ tasks, then for any $\delta \in (0, 1)$ with probability at least $1 - \delta$, for every solution $\theta_{1:t}$ of the continually learned $1 : t$ tasks in parameter space $\Theta$, i.e., $\theta_{1:t} \in \Theta$:*

$$\mathcal{E}_{\mathbb{D}_t}(\theta_{1:t}) < \hat{\mathcal{E}}^b_{\mathcal{D}_{1:t-1}}(\theta_{1:t}) + \frac{1}{2(t-1)} \sum_{k=1}^{t-1} \text{Div}(\mathbb{D}_k, \mathbb{D}_t) + \sqrt{\frac{d\ln(N_{1:t-1}/d) + \ln(1/\delta)}{N_{1:t-1}}}, \quad (S10)$$

$$\mathcal{E}_{\mathbb{D}_{1:t-1}}(\theta_{1:t}) < \hat{\mathcal{E}}^b_{\mathcal{D}_t}(\theta_{1:t}) + \frac{1}{2(t-1)} \sum_{k=1}^{t-1} \text{Div}(\mathbb{D}_t, \mathbb{D}_k) + \sqrt{\frac{d\ln(N_t/d) + \ln(1/\delta)}{N_t}}, \quad (S11)$$

*where $\text{Div}(\mathbb{D}_i, \mathbb{D}_j) := 2 \sup_{h \in \mathcal{H}} |\mathcal{P}_{\mathbb{D}_i}(I(h)) - \mathcal{P}_{\mathbb{D}_j}(I(h))|$ is the $\mathcal{H}$-divergence for $\mathbb{D}_i$ and $\mathbb{D}_j$ ($I(h)$ is the characteristic function). $N_{1:t-1} = \sum_{k=1}^{t-1} N_k$ is the total number of training samples over all old tasks.*

Below is an important definition and three critical lemmas for the proof of Proposition 3.

**Definition 1.** *(Based on Definition 1 of [2]) Given two distributions, $\mathbb{T}$ and $\mathbb{S}$, let $\mathcal{H}$ be a hypothesis class on input space $\mathcal{X}$ and denote by $I(h)$ the set for which $h \in \mathcal{H}$ is the characteristic function, i.e., $x \in I(h) \Leftrightarrow h(x) = 1$. The $\mathcal{H}$-divergence between $\mathbb{T}$ and $\mathbb{S}$ is*

$$\mathrm{Div}(\mathbb{T}, \mathbb{S}) = 2 \sup_{h \in \mathcal{H}} |\mathcal{P}_{\mathbb{T}}(I(h)) - \mathcal{P}_{\mathbb{S}}(I(h))|. \tag{S12}$$

**Lemma 1.** *Let $\mathbb{S} = \{\mathbb{S}_i\}_{i=1}^s$ be a set of source distributions and $\mathbb{T}$ be the target distribution. The $\mathcal{H}$-divergence between $\{\mathbb{S}_i\}_{i=1}^s$ and $\mathbb{T}$ is bounded as follows:*

$$\mathrm{Div}(\mathbb{S}, \mathbb{T}) \leq \frac{1}{s} \sum_{i=1}^s \mathrm{Div}(\mathbb{S}_i, \mathbb{T}). \tag{S13}$$

*Proof.* From the definition of $\mathcal{H}$-divergence,

$$
\begin{aligned}
\mathrm{Div}(\mathbb{S}, \mathbb{T}) &= 2 \sup_{h \in \mathcal{H}} |\mathcal{P}_{\mathbb{S}}(I(h)) - \mathcal{P}_{\mathbb{T}}(I(h))| \\
&= 2 \sup_{h \in \mathcal{H}} |\sum_{i=1}^s \frac{1}{s}(\mathcal{P}_{\mathbb{S}_i}(I(h)) - \mathcal{P}_{\mathbb{T}}(I(h)))| \\
&\leq 2 \sup_{h \in \mathcal{H}} \sum_{i=1}^s \frac{1}{s} |\mathcal{P}_{\mathbb{S}_i}(I(h)) - \mathcal{P}_{\mathbb{T}}(I(h))| \\
&\leq 2 \sum_{i=1}^s \frac{1}{s} \sup_{h \in \mathcal{H}} |\mathcal{P}_{\mathbb{S}_i}(I(h)) - \mathcal{P}_{\mathbb{T}}(I(h))| \\
&= \frac{1}{s} \sum_{i=1}^s \mathrm{Div}(\mathbb{S}_i, \mathbb{T}),
\end{aligned}
\tag{S14}
$$

where the first inequality derives from the triangle inequality and the second inequality derives from the additivity of the sup function. This finishes the proof.

**Lemma 2.** *Given two distributions $\mathbb{T}$ and $\mathbb{S}$, the difference between the population loss with $\mathbb{T}$ and $\mathbb{S}$ is bounded by the divergence between $\mathbb{T}$ and $\mathbb{S}$:*

$$|\mathcal{E}_{\mathbb{T}}(f_1, h_1) - \mathcal{E}_{\mathbb{S}}(f_1, h_1)| \leq \frac{1}{2} \mathrm{Div}(\mathbb{T}, \mathbb{S}), \tag{S15}$$

*where $\mathrm{Div}(\mathbb{T}, \mathbb{S}) := 2 \sup_{h \in \mathcal{H}} |\mathcal{P}_{\mathbb{T}}(I(h)) - \mathcal{P}_{\mathbb{S}}(I(h))|$ denotes the $\mathcal{H}$-divergence for the distributions $\mathbb{T}$ and $\mathbb{S}$ ($I(h)$ is the characteristic function).*

*Proof.* From the definition of $\mathcal{H}$-divergence,

$$
\begin{aligned}
\mathrm{Div}(\mathbb{T}, \mathbb{S}) &= 2 \sup_{h \in \mathcal{H}} |\mathcal{P}_{\mathbb{T}}(I(h)) - \mathcal{P}_{\mathbb{S}}(I(h))| \\
&= 2 \sup_{f_1, h_1 \in \mathcal{H}} |\mathcal{P}_{(x,y) \sim \mathbb{T}}[f_1(x) \neq h_1(x)] - \mathcal{P}_{(x,y) \sim \mathbb{S}}[f_1(x) \neq h_1(x)]| \\
&= 2 \sup_{f_1, h_1 \in \mathcal{H}} |\mathbb{E}_{(x,y) \sim \mathbb{T}}[\ell(f_1(x), h_1(x))] - \mathbb{E}_{(x,y) \sim \mathbb{S}}[\ell(f_1(x), h_1(x))]| \\
&= 2 \sup_{f_1, h_1 \in \mathcal{H}} |\mathcal{E}_{\mathbb{T}}(f_1, h_1) - \mathcal{E}_{\mathbb{S}}(f_1, h_1)| \\
&\geq 2|\mathcal{E}_{\mathbb{T}}(f_1, h_1) - \mathcal{E}_{\mathbb{S}}(f_1, h_1)|.
\end{aligned}
\tag{S16}
$$

It completes the proof of Lemma 2.

**Lemma 3.** *Let* $\Theta$ *be a cover of a parameter space with VC dimension* $d$. *Then, for any* $\delta \in (0,1)$ *with probability at least* $1 - \delta$, *for any* $\theta \in \Theta$,

$$|\mathcal{E}_{\mathbb{D}}(\theta) - \hat{\mathcal{E}}_D^b(\theta)| \leq \sqrt{\frac{d[\ln(N/d)] + \ln(1/\delta)}{2N}}, \tag{S17}$$

*where* $\hat{\mathcal{E}}_D^b(\theta)$ *is the robust empirical risk over the training set* $D$, $N$ *is the number of training samples, and* $b$ *is the radius around* $\theta$.

*Proof.* For the distribution $\mathbb{D}$, we have

$$\mathcal{P}(|\mathcal{E}_{\mathbb{D}}(\theta) - \hat{\mathcal{E}}_D(\theta)| \geq \epsilon) \leq 2m_{\Theta}(N)\exp(-2N\epsilon^2), \tag{S18}$$

where $m_{\Theta}(N)$ is the amount of all possible prediction results for $N$ samples, which implies the model complexity in the parameter space $\Theta$. We set $m_{\Theta}(N) = \frac{1}{2}\left(\frac{N}{d}\right)^d$ in our model, and assume a confidence bound $\epsilon = \sqrt{\frac{d[\ln(N/d)] + \ln(1/\delta)}{2N}}$. Then we get

$$\mathcal{P}(|\mathcal{E}_{\mathbb{D}}(\theta) - \hat{\mathcal{E}}_D(\theta)| \geq \epsilon) \leq \left(\frac{N}{d}\right)^d \exp(-2N\epsilon^2) = \delta. \tag{S19}$$

Hence, the inequality $|\mathcal{E}_{\mathbb{D}}(\theta) - \hat{\mathcal{E}}_D(\theta)| \leq \epsilon$ holds with probability at least $1 - \delta$. Further, based on the fact that $\hat{\mathcal{E}}_D^b(\theta) \geq \hat{\mathcal{E}}_D(\theta)$, we have

$$|\mathcal{E}_{\mathbb{D}}(\theta) - \hat{\mathcal{E}}_D^b(\theta)| \leq |\mathcal{E}_{\mathbb{D}}(\theta) - \hat{\mathcal{E}}_D(\theta)| \leq \epsilon. \tag{S20}$$

It completes the proof of Lemma 3.

*Proof of Proposition 3.* If we continually learn tasks $1:t$ that follow a set of distributions $\mathbb{D}_1, \cdots, \mathbb{D}_t$, then a solution $\theta_{1:t}$ can be obtained. Let $\theta_t$ denote a solution obtained over the distribution $\mathbb{D}_t$ only, and $\theta_{1:t-1}$ be a solution obtained over the distributions $\mathbb{D}_1, \cdots, \mathbb{D}_{t-1}$. Then we have

$$\begin{aligned}
\mathcal{E}_{\mathbb{D}_t}(\theta_{1:t-1}) &\leq \mathcal{E}_{\mathbb{D}_{1:t-1}}(\theta_{1:t-1}) + \frac{1}{2}\text{Div}(\mathbb{D}_{1:t-1}, \mathbb{D}_t) \\
&\leq \hat{\mathcal{E}}_{D_{1:t-1}}^b(\theta_{1:t-1}) + \frac{1}{2}\text{Div}(\mathbb{D}_{1:t-1}, \mathbb{D}_t) + \sqrt{\frac{d[\ln(N_{1:t-1}/d)] + \ln(1/\delta)}{2N_{1:t-1}}} \\
&\leq \hat{\mathcal{E}}_{D_{1:t-1}}^b(\theta_{1:t-1}) + \frac{1}{2(t-1)}\sum_{k=1}^{t-1}\text{Div}(\mathbb{D}_k, \mathbb{D}_t) + \sqrt{\frac{d[\ln(N_{1:t-1}/d)] + \ln(1/\delta)}{2N_{1:t-1}}} \\
&\leq \hat{\mathcal{E}}_{D_{1:t-1}}^b(\theta_{1:t-1}) + \frac{1}{2(t-1)}\sum_{k=1}^{t-1}\text{Div}(\mathbb{D}_k, \mathbb{D}_t) + \sqrt{\frac{d[\ln(N_{1:t-1}/d)] + \ln(1/\delta)}{N_{1:t-1}}},
\end{aligned} \tag{S21}$$

where the first three inequalities are from Lemma 2, Lemma 3 and Lemma 1, respectively. $\mathbb{D}_{1:t-1} := \{\mathbb{D}_k\}_{k=1}^{t-1}$ and we rewrite a mixture of all the $t-1$ distributions as $\mathbb{D}_{1:t-1} := \frac{1}{t-1}\sum_{k=1}^{t-1}\mathbb{D}_k$ using convex combination. $N_{1:t-1} = \sum_{k=1}^{t-1} N_k$ is the total number of training samples over all old tasks.

Further, we have

$$\begin{aligned}
&\mathcal{E}_{\mathbb{D}_t}(\theta_{1:t}) < \mathcal{E}_{\mathbb{D}_t}(\theta_{1:t-1}) \\
&\leq \hat{\mathcal{E}}_{D_{1:t-1}}^b(\theta_{1:t-1}) + \frac{1}{2(t-1)}\sum_{k=1}^{t-1}\text{Div}(\mathbb{D}_k,\mathbb{D}_t) + \sqrt{\frac{d[\ln(N_{1:t-1}/d)]+\ln(1/\delta)}{N_{1:t-1}}} \\
&\leq \hat{\mathcal{E}}_{D_{1:t-1}}^b(\theta_{1:t}) + \frac{1}{2(t-1)}\sum_{k=1}^{t-1}\text{Div}(\mathbb{D}_k,\mathbb{D}_t) + \sqrt{\frac{d[\ln(N_{1:t-1}/d)]+\ln(1/\delta)}{N_{1:t-1}}}.
\end{aligned} \tag{S22}$$

Likewise, we get

$$\begin{aligned}
\mathcal{E}_{\mathbb{D}_{1:t-1}}(\theta_t) &\leq \mathcal{E}_{\mathbb{D}_t}(\theta_t) + \frac{1}{2}\text{Div}(\mathbb{D}_t,\mathbb{D}_{1:t-1}) \\
&\leq \hat{\mathcal{E}}_{D_t}^b(\theta_t) + \frac{1}{2}\text{Div}(\mathbb{D}_t,\mathbb{D}_{1:t-1}) + \sqrt{\frac{d[\ln(N_t/d)]+\ln(1/\delta)}{2N_t}} \\
&\leq \hat{\mathcal{E}}_{D_t}^b(\theta_t) + \frac{1}{2(t-1)}\sum_{k=1}^{t-1}\text{Div}(\mathbb{D}_t,\mathbb{D}_k) + \sqrt{\frac{d[\ln(N_t/d)]+\ln(1/\delta)}{2N_t}} \\
&\leq \hat{\mathcal{E}}_{D_t}^b(\theta_t) + \frac{1}{2(t-1)}\sum_{k=1}^{t-1}\text{Div}(\mathbb{D}_t,\mathbb{D}_k) + \sqrt{\frac{d[\ln(N_t/d)]+\ln(1/\delta)}{N_t}}.
\end{aligned} \tag{S23}$$

Further, we have

$$\begin{aligned}
&\mathcal{E}_{\mathbb{D}_{1:t-1}}(\theta_{1:t}) < \mathcal{E}_{\mathbb{D}_{1:t-1}}(\theta_t) \\
&\leq \hat{\mathcal{E}}_{D_t}^b(\theta_t) + \frac{1}{2(t-1)}\sum_{k=1}^{t-1}\text{Div}(\mathbb{D}_t,\mathbb{D}_k) + \sqrt{\frac{d[\ln(N_t/d)]+\ln(1/\delta)}{N_t}} \\
&\leq \hat{\mathcal{E}}_{D_t}^b(\theta_{1:t}) + \frac{1}{2(t-1)}\sum_{k=1}^{t-1}\text{Div}(\mathbb{D}_t,\mathbb{D}_k) + \sqrt{\frac{d[\ln(N_t/d)]+\ln(1/\delta)}{N_t}},
\end{aligned} \tag{S24}$$

where $N_t$ is the number of training samples following the distribution $\mathbb{D}_t$.

Combining all the inequalities above finishes the proof of Proposition 3.

**Proposition 4.** *Let $\hat{\theta}_{1:t}^b$ denote the optimal solution of the continually learned $1:t$ tasks by robust empirical risk minimization over the current task, i.e., $\hat{\theta}_{1:t}^b = \arg\min_{\theta\in\Theta}\hat{\mathcal{E}}_{D_t}^b(\theta)$, where $\Theta$ denotes a cover of a parameter space with VC dimension $d$. Then, for any $\delta \in (0,1)$ with probability at least $1-\delta$:*

$$\mathcal{E}_{\mathbb{D}_t}(\hat{\theta}_{1:t}^b) - \min_{\theta\in\Theta}\mathcal{E}_{\mathbb{D}_t}(\theta) \leq \min_{\theta\in\Theta}\hat{\mathcal{E}}_{D_{1:t-1}}^b(\theta) - \min_{\theta\in\Theta}\hat{\mathcal{E}}_{D_{1:t-1}}(\theta) + \frac{1}{t-1}\sum_{k=1}^{t-1}\text{Div}(\mathbb{D}_k,\mathbb{D}_t) + C_1, \tag{S25}$$

$$\mathcal{E}_{\mathbb{D}_{1:t-1}}(\hat{\theta}_{1:t}^b) - \min_{\theta\in\Theta}\mathcal{E}_{\mathbb{D}_{1:t-1}}(\theta) \leq \min_{\theta\in\Theta}\hat{\mathcal{E}}_{D_t}^b(\theta) - \min_{\theta\in\Theta}\hat{\mathcal{E}}_{D_t}(\theta) + \frac{1}{t-1}\sum_{k=1}^{t-1}\text{Div}(\mathbb{D}_t,\mathbb{D}_k) + C_2, \tag{S26}$$

*where $C_1 = 2\sqrt{\frac{d\ln(N_{1:t-1}/d)+\ln(2/\delta)}{N_{1:t-1}}}$ and $C_2 = 2\sqrt{\frac{d\ln(N_t/d)+\ln(2/\delta)}{N_t}}$. $\text{Div}(\mathbb{D}_i,\mathbb{D}_j) := 2\sup_{h\in\mathcal{H}}|\mathcal{P}_{\mathbb{D}_i}(I(h)) - \mathcal{P}_{\mathbb{D}_j}(I(h))|$ is the $\mathcal{H}$-divergence for the distributions $\mathbb{D}_i$ and $\mathbb{D}_j$ ($I(h)$ is the characteristic function).*

*Proof of Proposition 4.* Let $\hat{\theta}^b_{1:t}$ denote the optimal solution of the continually learned $1:t$ tasks by robust empirical risk minimization over the current task, i.e., $\hat{\theta}^b_{1:t} = \arg\min_{\theta \in \Theta} \hat{\mathcal{E}}^b_{D_t}(\theta)$, where $\Theta$ denotes a cover of a parameter space with VC dimension $d$. Likewise, let $\hat{\theta}^b_t$ be the optimal solution by robust empirical risk minimization over the distribution $\mathbb{D}_t$ only, and $\hat{\theta}^b_{1:t-1}$ over the distributions $\mathbb{D}_1, \cdots, \mathbb{D}_{t-1}$. That is, $\hat{\theta}^b_t = \arg\min_\theta \hat{\mathcal{E}}^b_{D_t}(\theta)$ and $\hat{\theta}^b_{1:t-1} = \arg\min_\theta \hat{\mathcal{E}}^b_{D_{1:t-1}}(\theta)$.

Then, let $\theta_t$ be the optimal solution over the distribution $\mathbb{D}_t$ only, i.e., $\theta_t = \arg\min_\theta \mathcal{E}_{\mathbb{D}_t}(\theta)$. From Lemma 3, the following inequality holds with probability at least $1 - \frac{\delta}{2}$,

$$|\mathcal{E}_{\mathbb{D}_{1:t-1}}(\theta_t) - \hat{\mathcal{E}}_{D_{1:t-1}}(\theta_t)| \leq \sqrt{\frac{d\ln(N_{1:t-1}/d) + \ln(2/\delta)}{2N_{1:t-1}}} \leq \sqrt{\frac{d\ln(N_{1:t-1}/d) + \ln(2/\delta)}{N_{1:t-1}}}, \quad (S27)$$

where $N_{1:t-1} = \sum_{k=1}^{t-1} N_k$ is the total number of training samples over all old tasks.

Then, we have

$$\begin{aligned}
\min_{\theta \in \Theta} &\hat{\mathcal{E}}_{D_{1:t-1}}(\theta) \leq \hat{\mathcal{E}}_{D_{1:t-1}}(\theta_t) \\
&\leq \mathcal{E}_{\mathbb{D}_{1:t-1}}(\theta_t) + \sqrt{\frac{d\ln(N_{1:t-1}/d) + \ln(2/\delta)}{N_{1:t-1}}} \\
&\leq \mathcal{E}_{\mathbb{D}_t}(\theta_t) + \frac{1}{2}\text{Div}(\mathbb{D}_{1:t-1}, \mathbb{D}_t) + \sqrt{\frac{d\ln(N_{1:t-1}/d) + \ln(2/\delta)}{N_{1:t-1}}} \\
&= \min_{\theta \in \Theta} \mathcal{E}_{\mathbb{D}_t}(\theta) + \frac{1}{2}\text{Div}(\mathbb{D}_{1:t-1}, \mathbb{D}_t) + \sqrt{\frac{d\ln(N_{1:t-1}/d) + \ln(2/\delta)}{N_{1:t-1}}} \\
&\leq \min_{\theta \in \Theta} \mathcal{E}_{\mathbb{D}_t}(\theta) + \frac{1}{2(t-1)}\sum_{k=1}^{t-1}\text{Div}(\mathbb{D}_k, \mathbb{D}_t) + \sqrt{\frac{d\ln(N_{1:t-1}/d) + \ln(2/\delta)}{N_{1:t-1}}},
\end{aligned} \quad (S28)$$

where the third inequality holds from Lemma 2, and the final inequality is from Lemma 1.

From Proposition 3, the following inequality holds with probability at least $1 - \frac{\delta}{2}$,

$$\mathcal{E}_{\mathbb{D}_t}(\hat{\theta}^b_{1:t-1}) < \hat{\mathcal{E}}^b_{D_{1:t-1}}(\hat{\theta}^b_{1:t-1}) + \frac{1}{2(t-1)}\sum_{k=1}^{t-1}\text{Div}(\mathbb{D}_k, \mathbb{D}_t) + \sqrt{\frac{d\ln(N_{1:t-1}/d) + \ln(2/\delta)}{N_{1:t-1}}}. \quad (S29)$$

Combining Eq. (S28) and Eq. (S29), we get

$$\begin{aligned}
\mathcal{E}_{\mathbb{D}_t}(\hat{\theta}^b_{1:t}) - \min_{\theta \in \Theta} &\mathcal{E}_{\mathbb{D}_t}(\theta) \leq \mathcal{E}_{\mathbb{D}_t}(\hat{\theta}^b_{1:t-1}) - \min_{\theta \in \Theta} \mathcal{E}_{\mathbb{D}_t}(\theta) \\
&\leq \hat{\mathcal{E}}^b_{D_{1:t-1}}(\hat{\theta}^b_{1:t-1}) - \min_{\theta \in \Theta} \hat{\mathcal{E}}_{D_{1:t-1}}(\theta) + \frac{1}{t-1}\sum_{k=1}^{t-1}\text{Div}(\mathbb{D}_k, \mathbb{D}_t) + 2\sqrt{\frac{d\ln(N_{1:t-1}/d) + \ln(2/\delta)}{N_{1:t-1}}} \\
&= \min_{\theta \in \Theta} \hat{\mathcal{E}}^b_{D_{1:t-1}}(\theta) - \min_{\theta \in \Theta} \hat{\mathcal{E}}_{D_{1:t-1}}(\theta) + \frac{1}{t-1}\sum_{k=1}^{t-1}\text{Div}(\mathbb{D}_k, \mathbb{D}_t) + 2\sqrt{\frac{d\ln(N_{1:t-1}/d) + \ln(2/\delta)}{N_{1:t-1}}}.
\end{aligned} \quad (S30)$$

This completes the first part of Proposition 4.

Similarly, let $\theta_{1:t}$ be the optimal solution over the distribution $\mathbb{D}_{1:t-1}$ only, i.e., $\theta_{1:t-1} = \arg\min_\theta \mathcal{E}_{\mathbb{D}_{1:t-1}}(\theta)$. From Lemma 3, the following inequality holds with probability at least $1 - \frac{\delta}{2}$,

$$|\mathcal{E}_{\mathbb{D}_t}(\theta_{1:t-1}) - \hat{\mathcal{E}}_{D_t}(\theta_{1:t-1})| \leq \sqrt{\frac{d\ln(N_t/d) + \ln(2/\delta)}{2N_t}} \leq \sqrt{\frac{d\ln(N_t/d) + \ln(2/\delta)}{N_t}}, \quad (S31)$$

where $N_t$ is the number of training samples following the distribution $\mathbb{D}_t$.

Then, we have

$$
\begin{aligned}
\min_{\theta \in \Theta} \hat{\mathcal{E}}_{D_t}(\theta) &\leq \hat{\mathcal{E}}_{D_t}(\theta_{1:t-1}) \\
&\leq \mathcal{E}_{\mathbb{D}_t}(\theta_{1:t-1}) + \sqrt{\frac{d \ln(N_t/d) + \ln(2/\delta)}{N_t}} \\
&\leq \mathcal{E}_{\mathbb{D}_{1:t-1}}(\theta_{1:t-1}) + \frac{1}{2}\mathrm{Div}(\mathbb{D}_t, \mathbb{D}_{1:t-1}) + \sqrt{\frac{d \ln(N_t/d) + \ln(2/\delta)}{N_t}} \\
&= \min_{\theta \in \Theta} \mathcal{E}_{\mathbb{D}_{1:t-1}}(\theta) + \frac{1}{2}\mathrm{Div}(\mathbb{D}_t, \mathbb{D}_{1:t-1}) + \sqrt{\frac{d \ln(N_t/d) + \ln(2/\delta)}{N_t}} \\
&\leq \min_{\theta \in \Theta} \mathcal{E}_{\mathbb{D}_{1:t-1}}(\theta) + \frac{1}{2(t-1)}\sum_{k=1}^{t-1}\mathrm{Div}(\mathbb{D}_t, \mathbb{D}_k) + \sqrt{\frac{d \ln(N_t/d) + \ln(2/\delta)}{N_t}},
\end{aligned}
\tag{S32}
$$

where the third inequality holds from Lemma 2, and the final inequality is from Lemma 1.

From Proposition 3, the following inequality holds with probability at least $1 - \frac{\delta}{2}$,

$$
\mathcal{E}_{\mathbb{D}_{1:t-1}}(\hat{\theta}_t^b) < \hat{\mathcal{E}}_{D_t}^b(\hat{\theta}_t^b) + \frac{1}{2(t-1)}\sum_{k=1}^{t-1}\mathrm{Div}(\mathbb{D}_t, \mathbb{D}_k) + \sqrt{\frac{d \ln(N_t/d) + \ln(2/\delta)}{N_t}}.
\tag{S33}
$$

Combining Eq. (S32) and Eq. (S33), we get

$$
\begin{aligned}
\mathcal{E}_{\mathbb{D}_{1:t-1}}(\hat{\theta}_{1:t}^b) - \min_{\theta \in \Theta} \mathcal{E}_{\mathbb{D}_{1:t-1}}(\theta) &\leq \mathcal{E}_{\mathbb{D}_{1:t-1}}(\hat{\theta}_t^b) - \min_{\theta \in \Theta} \mathcal{E}_{\mathbb{D}_{1:t-1}}(\theta) \\
&\leq \hat{\mathcal{E}}_{D_t}^b(\hat{\theta}_t^b) - \min_{\theta \in \Theta} \hat{\mathcal{E}}_{D_t}(\theta) + \frac{1}{t-1}\sum_{k=1}^{t-1}\mathrm{Div}(\mathbb{D}_t, \mathbb{D}_k) + 2\sqrt{\frac{d \ln(N_t/d) + \ln(2/\delta)}{N_t}} \\
&= \min_{\theta \in \Theta} \hat{\mathcal{E}}_{D_t}^b(\theta) - \min_{\theta \in \Theta} \hat{\mathcal{E}}_{D_t}(\theta) + \frac{1}{t-1}\sum_{k=1}^{t-1}\mathrm{Div}(\mathbb{D}_t, \mathbb{D}_k) + 2\sqrt{\frac{d \ln(N_t/d) + \ln(2/\delta)}{N_t}}.
\end{aligned}
\tag{S34}
$$

This completes the second part of Proposition 4.

Below is one critical lemma for the proof of Proposition 1.

**Lemma 4.** *Let $\{\Theta_i \in \mathbb{R}^{M_i}\}_{i=1}^{K}$ be a set of $K$ parameter spaces ($K \geq 1$ in general), $d_i$ be a VC dimension of $\Theta_i$, and $\Theta = \cup_{i=1}^{K}\Theta_i$ with VC dimension $d$. Let $\theta_i = \arg\max_{\theta \in \Theta_i} \mathcal{E}_{\mathbb{D}}(\theta)$ be a local maximum in the $i$-th parameter space (i.e., $i$-th ball). Then, for any $\delta \in (0, 1)$ with probability at least $1 - \delta$, for any $\theta \in \Theta$,*

$$
|\mathcal{E}_{\mathbb{D}}(\theta) - \hat{\mathcal{E}}_D^b(\theta)| \leq \max_{i \in [1,K]} \sqrt{\frac{d_i \ln(N/d_i) + \ln(K/\delta)}{2N}},
\tag{S35}
$$

*where $\hat{\mathcal{E}}_D^b(\theta)$ is the robust empirical risk over the training set $D$, $N$ is the number of training samples following the distribution $\mathbb{D}$, and $b$ is the radius around $\theta$.*

*Proof.* For the distribution $\mathbb{D}$ with training set $D$, we have

$$
\mathcal{P}\left(\max_{i \in [1,K]} |\mathcal{E}_{\mathbb{D}}(\theta_i) - \hat{\mathcal{E}}_D(\theta_i)| \geq \epsilon\right) \leq \sum_{i=1}^{K} \mathcal{P}\left(|\mathcal{E}_{\mathbb{D}}(\theta_i) - \hat{\mathcal{E}}_D(\theta_i)| \geq \epsilon\right) \leq \sum_{i=1}^{K} 2m_{\Theta_i}(N)\exp(-2N\epsilon^2),
\tag{S36}
$$

where $m_{\Theta_i}(N)$ is the amount of all possible prediction results for $N$ samples, which implies the model complexity in the parameter space $\Theta_i$. We set $m_{\Theta_i}(N) = \frac{1}{2}\left(\frac{N}{d_i}\right)^{d_i}$ in our model, and assume a confidence bound $\epsilon_i = \sqrt{\frac{d_i[\ln(N/d_i)] + \ln(K/\delta)}{2N}}$, and $\epsilon = \max_{i \in [1,K]} \epsilon_i$. In addition, to make the continual learners well capture the parameter space $\Theta$, we consider $K = \left\lceil \left(\frac{\sup_{\theta, \theta' \in \Theta} \|\theta - \theta'\|_2}{b}\right)^M \right\rceil$, where $M = \sum_{i=1}^{K} M_i$ is the total dimension budget of $\Theta$. Therefore, the trade-off between learner number (i.e., $K$) and width (i.e., $M_i$) is independent of the training data distribution $\mathbb{D}$.

Then we get

$$\begin{aligned}
\mathcal{P}\left(\max_{i \in [1,K]} |\mathcal{E}_{\mathbb{D}}(\theta_i) - \hat{\mathcal{E}}_D(\theta_i)| \geq \epsilon\right) &\leq \sum_{i=1}^{K} 2m_{\Theta_i}(N)\exp(-2N\epsilon^2) \\
&= \sum_{i=1}^{K} \left(\frac{N}{d_i}\right)^{d_i}\exp(-2N\epsilon^2) \\
&\leq \sum_{i=1}^{K} \left(\frac{N}{d_i}\right)^{d_i}\exp(-2N\epsilon_i^2) \\
&= \sum_{i=1}^{K} \frac{\delta}{K} = \delta.
\end{aligned} \tag{S37}$$

Hence, the inequality $|\mathcal{E}_{\mathbb{D}}(\theta) - \hat{\mathcal{E}}_D(\theta)| \leq \epsilon$ holds with probability at least $1 - \delta$. Further, based on the fact that $\hat{\mathcal{E}}_D^b(\theta) \geq \hat{\mathcal{E}}_D(\theta)$, we have

$$|\mathcal{E}_{\mathbb{D}}(\theta) - \hat{\mathcal{E}}_D^b(\theta)| \leq |\mathcal{E}_{\mathbb{D}}(\theta) - \hat{\mathcal{E}}_D(\theta)| \leq \epsilon. \tag{S38}$$

It completes the proof of Lemma 4.

*Proof of Proposition 1.* Let $\{\Theta_i \in \mathbb{R}^{M_i}\}_{i=1}^{K}$ be a set of $K$ parameter spaces ($K \geq 1$ in general), $d_i$ be a VC dimension of $\Theta_i$, and $\Theta = \cup_{i=1}^{K}\Theta_i$ with VC dimension $d$. Let $\hat{\theta}_{1:t}^b$ denote the optimal solution of the continually learned $1 : t$ tasks by robust empirical risk minimization over the current task, i.e., $\hat{\theta}_{1:t}^b = \arg\min_{\theta \in \Theta} \hat{\mathcal{E}}_{D_t}^b(\theta)$, where $\Theta$ denotes a cover of a parameter space with VC dimension $d$. Likewise, let $\hat{\theta}_t^b$ be the optimal solution by robust empirical risk minimization over the distribution $\mathbb{D}_t$ only, and $\hat{\theta}_{1:t-1}^b$ is over the distributions $\mathbb{D}_1, \cdots, \mathbb{D}_{t-1}$. That is, $\hat{\theta}_t^b = \arg\min_{\theta} \hat{\mathcal{E}}_{D_t}^b(\theta)$ and $\hat{\theta}_{1:t-1}^b = \arg\min_{\theta} \hat{\mathcal{E}}_{D_{1:t-1}}^b(\theta)$.

Then, let $\theta_t$ be the optimal solution over the distribution $\mathbb{D}_t$ only, i.e., $\theta_t = \arg\min_{\theta} \mathcal{E}_{\mathbb{D}_t}(\theta)$. From Lemma 3 and Proposition 4, the following inequality holds with probability at least $1 - \frac{\delta}{2}$,

$$|\mathcal{E}_{\mathbb{D}_t}(\theta_{1:t-1}) - \hat{\mathcal{E}}_{D_t}(\theta_{1:t-1})| \leq \sqrt{\frac{d\ln(N_t/d) + \ln(2/\delta)}{N_t}}, \tag{S39}$$

where $N_{1:t-1} = \sum_{k=1}^{t-1} N_k$ is the total number of training samples over all old tasks. Then we have

$$\min_{\theta \in \Theta} \hat{\mathcal{E}}_{D_{1:t-1}}(\theta) \leq \min_{\theta \in \Theta} \mathcal{E}_{\mathbb{D}_t}(\theta) + \frac{1}{2(t-1)}\sum_{k=1}^{t-1} \text{Div}(\mathbb{D}_k, \mathbb{D}_t) + \sqrt{\frac{d\ln(N_t/d) + \ln(2/\delta)}{N_t}}. \tag{S40}$$

From Proposition 3 and Lemma 4, the following inequality holds with probability at least $1 - \frac{\delta}{2}$,

$$\mathcal{E}_{\mathbb{D}_t}(\hat{\theta}^b_{1:t-1}) < \hat{\mathcal{E}}^b_{D_{1:t-1}}(\hat{\theta}^b_{1:t-1}) + \frac{1}{2(t-1)}\sum_{k=1}^{t-1}\mathrm{Div}(\mathbb{D}_k,\mathbb{D}_t) + \max_{i\in[1,K]}\sqrt{\frac{d_i\ln(N_{1:t-1}/d_i) + \ln(2K/\delta)}{2N_{1:t-1}}}. \tag{S41}$$

Combining Eq. (S40) and Eq. (S41), we get

$$\begin{aligned}
\mathcal{E}_{\mathbb{D}_t}(\hat{\theta}^b_{1:t}) - \min_{\theta\in\Theta}\mathcal{E}_{\mathbb{D}_t}(\theta) &\leq \mathcal{E}_{\mathbb{D}_t}(\hat{\theta}^b_{1:t-1}) - \min_{\theta\in\Theta}\mathcal{E}_{\mathbb{D}_t}(\theta) \\
&\leq \hat{\mathcal{E}}^b_{D_{1:t-1}}(\hat{\theta}^b_{1:t-1}) - \min_{\theta\in\Theta}\hat{\mathcal{E}}_{D_{1:t-1}}(\theta) + \frac{1}{t-1}\sum_{k=1}^{t-1}\mathrm{Div}(\mathbb{D}_k,\mathbb{D}_t) \\
&\quad + \max_{i\in[1,K]}\sqrt{\frac{d_i\ln(N_{1:t-1}/d_i) + \ln(2K/\delta)}{2N_{1:t-1}}} + \sqrt{\frac{d\ln(N_{1:t-1}/d) + \ln(2/\delta)}{N_{1:t-1}}} \\
&= \min_{\theta\in\Theta}\hat{\mathcal{E}}^b_{D_{1:t-1}}(\theta) - \min_{\theta\in\Theta}\hat{\mathcal{E}}_{D_{1:t-1}}(\theta) + \frac{1}{t-1}\sum_{k=1}^{t-1}\mathrm{Div}(\mathbb{D}_k,\mathbb{D}_t) \\
&\quad + \max_{i\in[1,K]}\sqrt{\frac{d_i\ln(N_{1:t-1}/d_i) + \ln(2K/\delta)}{2N_{1:t-1}}} + \sqrt{\frac{d\ln(N_{1:t-1}/d) + \ln(2/\delta)}{N_{1:t-1}}}.
\end{aligned} \tag{S42}$$

This completes the first part of Proposition 1.

Similarly, let $\theta_{1:t}$ be the optimal solution over the distribution $\mathbb{D}_{1:t-1}$ only, i.e., $\theta_{1:t-1} = \arg\min_\theta \mathcal{E}_{\mathbb{D}_{1:t-1}}(\theta)$. From Lemma 3 and Proposition 4, the following inequality holds with probability at least $1 - \frac{\delta}{2}$,

$$|\mathcal{E}_{\mathbb{D}_t}(\theta_{1:t-1}) - \hat{\mathcal{E}}_{D_t}(\theta_{1:t-1})| \leq \sqrt{\frac{d\ln(N_t/d) + \ln(2/\delta)}{N_t}}, \tag{S43}$$

where $N_t$ is the number of training samples following the distribution $\mathbb{D}_t$. Then, we have

$$\min_{\theta\in\Theta}\hat{\mathcal{E}}_{D_t}(\theta) \leq \min_{\theta\in\Theta}\mathcal{E}_{\mathbb{D}_{1:t-1}}(\theta) + \frac{1}{2(t-1)}\sum_{k=1}^{t-1}\mathrm{Div}(\mathbb{D}_t,\mathbb{D}_k) + \sqrt{\frac{d\ln(N_t/d) + \ln(2/\delta)}{N_t}}. \tag{S44}$$

From Proposition 3 and Lemma 4, the following inequality holds with probability at least $1 - \frac{\delta}{2}$,

$$\mathcal{E}_{\mathbb{D}_{1:t-1}}(\hat{\theta}^b_t) < \hat{\mathcal{E}}^b_{D_t}(\hat{\theta}^b_t) + \frac{1}{2(t-1)}\sum_{k=1}^{t-1}\mathrm{Div}(\mathbb{D}_t,\mathbb{D}_k) + \max_{i\in[1,K]}\sqrt{\frac{d_i\ln(N_t/d_i) + \ln(2K/\delta)}{2N_t}}. \tag{S45}$$

Combining Eq. (S44) and Eq. (S45), we get

$$\begin{aligned}
\mathcal{E}_{\mathbb{D}_{1:t-1}}(\hat{\theta}^b_{1:t}) - \min_{\theta\in\Theta}\mathcal{E}_{\mathbb{D}_{1:t-1}}(\theta) &\leq \mathcal{E}_{\mathbb{D}_{1:t-1}}(\hat{\theta}^b_t) - \min_{\theta\in\Theta}\mathcal{E}_{\mathbb{D}_{1:t-1}}(\theta) \\
&\leq \hat{\mathcal{E}}^b_{D_t}(\hat{\theta}^b_t) - \min_{\theta\in\Theta}\hat{\mathcal{E}}_{D_t}(\theta) + \frac{1}{t-1}\sum_{k=1}^{t-1}\mathrm{Div}(\mathbb{D}_t,\mathbb{D}_k) \\
&\quad + \max_{i\in[1,K]}\sqrt{\frac{d_i\ln(N_t/d_i) + \ln(2K/\delta)}{2N_t}} + \sqrt{\frac{d\ln(N_t/d) + \ln(1/\delta)}{N_t}} \\
&= \min_{\theta\in\Theta}\hat{\mathcal{E}}^b_{D_t}(\theta) - \min_{\theta\in\Theta}\hat{\mathcal{E}}_{D_t}(\theta) + \frac{1}{t-1}\sum_{k=1}^{t-1}\mathrm{Div}(\mathbb{D}_t,\mathbb{D}_k) \\
&\quad + \max_{i\in[1,K]}\sqrt{\frac{d_i\ln(N_t/d_i) + \ln(2K/\delta)}{2N_t}} + \sqrt{\frac{d\ln(N_t/d) + \ln(1/\delta)}{N_t}}.
\end{aligned} \tag{S46}$$

This completes the second part of Proposition 1.

**Proof of Proposition 2:** *Proof of Proposition 2.* Below, we first state a more explicit generalization bound based on the PAC-Bayes theory in terms of Proposition 3. That is, if $\mathbb{D}_1, \cdots, \mathbb{D}_t$ are the distributions of the continually learned $1:t$ tasks, then for any $\delta \in (0,1)$ with probability at least $1-\delta$, for every solution $\theta_{1:t}$ of the continually learned $1:t$ tasks in parameter space $\Theta$ with a $L_2$ norm regularization, i.e., $\theta_{1:t} \in \Theta$, $\|\theta_{1:t}\|_2^2 \leq \rho$, and $\rho > 0$:

$$\mathcal{E}_{\mathbb{D}_t}(\theta_{1:t}) < \hat{\mathcal{E}}_{D_{1:t-1}}^b(\theta_{1:t}) + \sqrt{\frac{M \ln\left(1 + \frac{\|\theta_{1:t}\|_2^2}{b^2}\left(1 + \sqrt{\frac{\ln N_{1:t-1}}{M}}\right)^2\right) + 4\ln\frac{N_{1:t-1}}{\delta} + \tilde{O}(1)}{N_{1:t-1} - 1}}$$

$$\leq \hat{\mathcal{E}}_{D_{1:t-1}}^b(\theta_{1:t}) + \frac{1}{2(t-1)}\sum_{k=1}^{t-1}\text{Div}(\mathbb{D}_k, \mathbb{D}_t) + r_1(\|\theta_{1:t}\|_2^2/b^2),$$

$$(S47)$$

$$\mathcal{E}_{\mathbb{D}_{1:t-1}}(\theta_{1:t}) < \hat{\mathcal{E}}_{D_t}^b(\theta_{1:t}) + \sqrt{\frac{M \ln\left(1 + \frac{\|\theta_{1:t}\|_2^2}{b^2}\left(1 + \sqrt{\frac{\ln N_t}{M}}\right)^2\right) + 4\ln\frac{N_t}{\delta} + \tilde{O}(1)}{N_t - 1}}$$

$$\leq \hat{\mathcal{E}}_{D_t}^b(\theta_{1:t}) + \frac{1}{2(t-1)}\sum_{k=1}^{t-1}\text{Div}(\mathbb{D}_t, \mathbb{D}_k) + r_2(\|\theta_{1:t}\|_2^2/b^2),$$

$$(S48)$$

where

$$r_1(\|\theta_{1:t}\|_2^2/b^2) = \sqrt{\frac{M \ln\left(1 + \frac{\|\theta_{1:t}\|_2^2}{b^2}\left(1 + \sqrt{\frac{\ln N_{1:t-1}}{M}}\right)^2\right) + 4\ln\frac{N_{1:t-1}}{\delta}}{N_{1:t-1} - 1}},$$

$$(S49)$$

$$r_2(\|\theta_{1:t}\|_2^2/b^2) = \sqrt{\frac{M \ln\left(1 + \frac{\|\theta_{1:t}\|_2^2}{b^2}\left(1 + \sqrt{\frac{\ln N_t}{M}}\right)^2\right) + 4\ln\frac{N_t}{\delta}}{N_t - 1}}.$$

$$(S50)$$

The first inequalities in Eq. (S47) and Eq. (S48) are derived from Theorem 2 in [7], which hold by assuming $\mathcal{E}_{\mathbb{D}_t}(\theta_{1:t}) < \mathbb{E}_{\Delta_i \sim \mathcal{N}(0,b)}[\mathcal{E}_{\mathbb{D}_t}(\theta_{1:t} + \Delta)]$ and $\mathcal{E}_{\mathbb{D}_{1:t-1}}(\theta_{1:t}) < \mathbb{E}_{\Delta_i \sim \mathcal{N}(0,b)}[\mathcal{E}_{\mathbb{D}_{1:t-1}}(\theta_{1:t} + \Delta)]$.

With active forgetting taking the form of $L_2$ norm over parameters, we need to substitute the third term in Eq. (S10) and Eq. (S11) with the specific increasing function $r_1$ or $r_2$ as in Proposition 3. From Eq. (S47) and Eq. (S48), it can be concluded that minimizing the $L_2$ norm of $\theta_{1:t}$ (i.e., active forgetting) can indeed tighten the generalization bound. Likewise, we need to modify Proposition 4 with $C_1 = r_1(\|\hat{\theta}_{1:t}^b\|_2^2/b^2)$ and $C_2 = r_2(\|\hat{\theta}_{1:t}^b\|_2^2/b^2)$. Based on the improved Proposition 3 and Proposition 4, we obtain Proposition 2.

# 2  Implementation Detail

## 2.1  Network Architecture

Following the previous work [9, 3, 18, 19], we use a 6-layer CNN for S-CIFAR-100, R-CIFAR-100 and R-CIFAR-10/100, a 4-layer CNN for Omniglot, a regular AlexNet [11] for CUB-200-2011, Tiny-ImageNet and CORe50, and a 3-layer CNN for Atari reinforcement tasks. The network architectures are detailed in Supplementary Table 1-4 (the output head is not included). The last dense layer serves as the output weight for each learner in our approach. We provide empirical results in Supplementary Table 5 to determine an appropriate trade-off between learner number and width. It can be seen that the use of five learners ($K = 5$) with appropriate width indeed achieves outstanding performance.

**Supplementary Table 1  Network architecture for S-CIFAR-100, R-CIFAR-100 and R-CIFAR-10/100.** We set $nc = 32$ for the SCL (0.84M parameters) while $nc = 9$ for our approach ($K = 5$, 0.88M parameters).

| Layer | Channel | Kernel | Stride | Padding | Dropout |
|---|---|---|---|---|---|
| Input | 3 | | | | |
| Conv 1 | $nc$ | 3×3 | 1 | 1 | |
| Conv 2 | $nc$ | 3×3 | 1 | 1 | |
| MaxPool | | | 2 | 0 | 0.25 |
| Conv 3 | $2nc$ | 3×3 | 1 | 1 | |
| Conv 4 | $2nc$ | 3×3 | 1 | 1 | |
| MaxPool | | | 2 | 0 | 0.25 |
| Conv 5 | $4nc$ | 3×3 | 1 | 1 | |
| Conv 6 | $4nc$ | 3×3 | 1 | 1 | |
| MaxPool | | | 2 | 1 | 0.25 |
| Dense 1 | 256 | | | | |

**Supplementary Table 2  Network architecture for Omniglot.** We set $nc = 64$ for the SCL (1.8M parameters) while $nc = 28$ for our approach ($K = 5$, 1.8M parameters). *Dense 1 is not used in SCL.

| Layer | Channel | Kernel | Stride | Padding | Dropout |
|---|---|---|---|---|---|
| Input | 1 | | | | |
| Conv 1 | $nc$ | 3×3 | 1 | 0 | |
| Conv 2 | $nc$ | 3×3 | 1 | 0 | |
| MaxPool | | | 2 | 0 | |
| Conv 3 | $nc$ | 3×3 | 1 | 0 | |
| Conv 4 | $nc$ | 3×3 | 1 | 0 | |
| MaxPool | | | 2 | 0 | |
| *Dense 1 | $16nc$ | | | | |

**Supplementary Table 3 Network architecture for CUB-200-2011, Tiny-ImageNet and CORe50.** We set $nc = 64$ for the SCL (57.8M parameters) while $nc = 34$ for our approach ($K = 5$, 57.2M parameters).

| Layer | Channel | Kernel | Stride | Padding | Dropout |
|-------|---------|--------|--------|---------|---------|
| Input | 3 | | | | |
| Conv 1 | $nc$ | 11×11 | 4 | 2 | |
| MaxPool | | 3×3 | 2 | 0 | |
| Conv 2 | $3nc$ | 5×5 | 1 | 2 | |
| MaxPool | | 3×3 | 2 | 0 | |
| Conv 3 | $6nc$ | 3×3 | 1 | 1 | |
| Conv 4 | $4nc$ | 3×3 | 1 | 1 | |
| Conv 5 | $4nc$ | 3×3 | 1 | 1 | |
| MaxPool | | 3×3 | 2 | 0 | |
| Dense 1 | $64nc$ | | | | 0.5 |
| Dense 2 | $64nc$ | | | | 0.5 |

**Supplementary Table 4 Network architecture for Atari reinforcement tasks.** We set $nc = 128$ for the SCL (4.1M parameters) while $nc = 32$ for our approach ($K = 5$, 4.3M parameters).

| Layer | Channel | Kernel | Stride | Padding | Dropout |
|-------|---------|--------|--------|---------|---------|
| Input | 4 | | | | |
| Conv 1 | $nc$ | 8×8 | 4 | 0 | |
| ReLU | | | | | |
| Conv 2 | $nc$ | 4×4 | 2 | 0 | |
| ReLU | | | | | |
| Conv 3 | $2nc$ | 3×3 | 1 | 0 | |
| ReLU | | | | | |
| Flatten | | | | | |
| Dense 1 | $49nc$ | | | | |

**Supplementary Table 5 Number-width trade-off under similar parameter budgets.** We use EWC [10] as the baseline approach, and report the average accuracy (%) over 5 runs with different random seeds and task orders.

| Learner Number | 3 | 4 | 5 | 6 | 7 | 8 |
|----------------|-----|-----|-----|-----|-----|-----|
| S-CIFAR-100 | 55.97 | 56.84 | 57.43 | 57.69 | – | – |
| R-CIFAR-100 | 72.43 | 73.17 | 73.27 | 73.18 | – | – |
| R-CIFAR-10/100 | 80.74 | 80.94 | 81.21 | 81.06 | – | – |
| Omniglot | 86.65 | 89.13 | 88.71 | 88.68 | 87.30 | 86.83 |
| CUB-200-2011 | 46.08 | 47.15 | 47.28 | 48.43 | 48.78 | 48.33 |
| Tiny-ImageNet | 52.56 | 53.05 | 53.85 | 54.03 | 54.63 | 54.30 |

## 2.2 Training Regime

For S-CIFAR-100, R-CIFAR-100, R-CIFAR-10/100 and Omniglot, we use an Adam optimizer of initial learning rate 0.001 and train all approaches with batch size of 256 for 100 epochs. For CUB-200-2011 and Tiny-ImageNet, we use a SGD optimizer of initial learning rate 0.005 and momentum 0.9, and train all approaches with batch size of 64 for 40 epochs. For CORe50, we use a SGD optimizer of learning rate 0.003, and train all approaches with batch size of 32 for 40 epochs. For Atari reinforcement tasks, we adopt the same PPO algorithm [14] to learn each task, following the training regime of previous work [9, 3, 18]. We use an Adam optimizer with initial learning rate of 0.0003, and evaluate the normalized accumulated reward of all tasks ever seen for 30 times during the training of each task.

## 2.3 Hyperparameter

Here we report the grid of values evaluated in hyperparameter search for all approaches and **bold** the selected ones. The hyperparameters used for S-CIFAR-100, R-CIFAR-100, R-CIFAR-10/100, Omniglot, CUB-200-2011 and Tiny-ImageNet are summarized in Supplementary Table 6. The hyperparameters used for CORe50 include: $\lambda_{\text{SP}}(1, 10, \textbf{100}, 1000)$ for EWC [10], $\lambda_{\text{SP}}(0.001, \textbf{0.01}, 0.1, 1)$ for MAS [1], and $\lambda_{\text{AF}}(\text{1e-8}, \text{1e-7}, \textbf{1e-6}, \text{1e-5})$, $\gamma(0.001, 0.01, \textbf{0.1}, 1)$ for CAF. The hyperparameters used for Atari reinforcement tasks include: $\lambda_{\text{SP}}(10000, 50000, \textbf{100000}, 200000, 400000)$ for EWC [10], $\lambda_{\text{SP}}(1, 5, \textbf{10}, 20, 40, 100)$ for MAS [1], $\lambda_{\text{CPR}}(0.01, 0.02, \textbf{0.05}, 0.1, 1)$ for CPR [3], and $\lambda_{\text{AF}}(\text{1e-7}, \text{1e-6}, \textbf{1e-5}, \text{1e-4})$, $\gamma(0.001, \textbf{0.01}, 0.1, 1)$ for CAF.

**Supplementary Table 6 Hyperparameters for continual learning of visual classification tasks.** We make an extensive hyperparameter search for all approaches and apply the best ones (marked in bold) to perform actual experiments. Due to differences in scale and sensitivity, the grid of values is referenced to published papers and refined by empirical results.

| Methods | S-CIFAR-100 | R-CIFAR-100 | R-CIFAR-10/100 |
|---|---|---|---|
| EWC [16] | $\lambda_{ewc}$(10000, **25000**, 50000, 100000) | $\lambda_{ewc}$(5000, 10000, **25000**, 50000) | $\lambda_{ewc}$(5000, 10000, **25000**, 50000) |
| MAS [1] | $\lambda_{ref}$(4, 8, **16**, 32) | $\lambda_{ref}$(2, **4**, 8, 16) | $\lambda_{ref}$(**4**, 6, 8, 16) |
| SI [20] | $\lambda_{si}$(4, 6, 8, **10**, 16) | $\lambda_{si}$(4, 8, **10**, 12, 16) | $\lambda_{si}$(4, 6, **8**, 16) |
| AGS-CL [9] | $\lambda$(800, 1600, **3200**, 7000), $\mu$(2, 5, **10**, 20), $\rho$(0.1, 0.2, **0.3**, 0.5) | $\lambda$(800, **1600**, 3200, 7000), $\mu$(2, 5, **10**, 20), $\rho$(0.1, 0.2, **0.3**, 0.5) | $\lambda$(1600, 3200, **7000**, 10000), $\mu$(5, 10, **20**, 40), $\rho$(0.1, **0.2**, 0.3, 0.5) |
| P&C [15] | $\lambda_{ewc}$(20000, **50000**, 100000, 200000) | $\lambda_{ewc}$(10000, **20000**, 50000, 100000) | $\lambda_{ewc}$(10000, 20000, **50000**, 100000) |
| CPR-EWC [3] | $\lambda_{ewc}$(10000, **25000**, 50000, 100000), $\lambda_{cpr}$(0.1, 0.5, 1, **1.5**, 3) | $\lambda_{ewc}$(5000, **10000**, 25000, 50000), $\lambda_{cpr}$(0.1, 0.5, 1, **1.5**, 3) | $\lambda_{ewc}$(5000, 10000, **25000**, 50000), $\lambda_{cpr}$(0.1, 0.2, **0.5**, 1, 1.5) |
| CPR-MAS [3] | $\lambda_{ref}$(4, 8, **16**, 32), $\lambda_{cpr}$(0.1, 0.5, 1, **1.5**, 3) | $\lambda_{ref}$(2, **4**, 8, 16), $\lambda_{cpr}$(0.1, 0.5, 1, **1.5**, 3) | $\lambda_{ref}$(**4**, 6, 8, 16), $\lambda_{cpr}$(0.1, **0.2**, 0.5, 1, 1.5) |
| CAF-EWC | $\lambda_{ewc}$(**10000**, 25000), $\lambda_{af}$(1e-6, 1e-7, **1e-8**, 1e-9), $\gamma$(0.001, **0.01**, 0.02, 0.1) | $\lambda_{ewc}$(**5000**, 10000), $\lambda_{af}$(1e-6, 1e-7, **1e-8**, 1e-9), $\gamma$(0.001, 0.01, **0.02**, 0.1) | $\lambda_{ewc}$(**10000**, 25000), $\lambda_{af}$(1e-6, 1e-7, **1e-8**, 1e-9), $\gamma$(0.001, 0.01, **0.02**, 0.1) |
| CAF-MAS | $\lambda_{ref}$(**8**, 16), $\lambda_{af}$(1e-6, 1e-7, **1e-8**, 1e-9), $\gamma$(0.001, 0.01, **0.02**, 0.1) | $\lambda_{ref}$(2, **4**), $\lambda_{af}$(1e-6, 1e-7, **1e-8**, 1e-9), $\gamma$(0.001, 0.01, **0.02**, 0.1) | $\lambda_{ref}$(4, **8**), $\lambda_{af}$(1e-6, 1e-7, **1e-8**, 1e-9), $\gamma$(0.0001, 0.01, **0.02**, 0.1) |

| Methods | Omniglot | CUB-200-2011 | Tiny-ImageNet |
|---|---|---|---|
| EWC [16] | $\lambda_{ewc}$(1e5, 1e6, 2e6, **5e6**, 1e7) | $\lambda_{ewc}$(0.5, 1, **2**, 5, 10, 20) | $\lambda_{ewc}$(20, **40**, 160) |
| MAS [1] | $\lambda_{ref}$(2, **4**, 8, 16) | $\lambda_{ref}$(0.001, 0.005, **0.01**, 0.02, 0.1) | $\lambda_{ref}$(0.1, 0.2, **0.5**, 1, 2) |
| SI [20] | $\lambda_{si}$(4, 6, 8, **10**, 16) | $\lambda_{si}$(0.2, **0.4**, 0.8, 1.6) | $\lambda_{si}$(0.4, **0.8**, 1.6, 3.2) |
| AGS-CL [9] | $\lambda$(500, 800, 1000, 1600, 3200), $\mu$(2, 5, **7**, 10), $\rho$(0.1, 0.3, **0.5**, 0.7) | $\lambda$(0.0001, **0.001**, 0.01, 0.1, 1, 10, 100), $\mu$(**0.01**), $\rho$(0.05, **0.1**, 0.2, 0.5) | $\lambda$(0.01, **0.1**, 1, 10, 100), $\mu$(0.0001, **0.001**, 0.01, 0.1), $\rho$(0.05, **0.1**, 0.2, 0.5) |
| P&C [15] | $\lambda_{ewc}$(2e6, 5e6, **1e7**, 2e7) | $\lambda_{ewc}$(1, 2, 4, **10**, 20, 40) | $\lambda_{ewc}$(40, **80**, 160, 320) |
| CPR-EWC [3] | $\lambda_{ewc}$(1e5, 1e6, 2e6, **5e6**, 1e7), $\lambda_{cpr}$(0.1, **0.2**, 0.5, 1) | $\lambda_{ewc}$(0.5, 1, 2, 5, 10, 20), $\lambda_{cpr}$(0.1, 0.2, 0.5, 1, 2) | $\lambda_{ewc}$(20, **40**, 80, 160), $\lambda_{cpr}$(0.1, 0.2, **0.5**, 1) |
| CPR-MAS [3] | $\lambda_{ref}$(2, **4**, 8, 16), $\lambda_{cpr}$(0.1, 0.2, 0.5, 1) | $\lambda_{ref}$(0.001, 0.005, **0.01**, 0.02, 0.1), $\lambda_{cpr}$(0.1, 0.2, 0.5, 1) | $\lambda_{ref}$(0.1, 0.2, **0.5**, 1, 2), $\lambda_{cpr}$(0.1, 0.2, 0.5, 1, 2) |
| CAF-EWC | $\lambda_{ewc}$(**5e6**, 1e7), $\lambda_{af}$(1e-10, 1e-11, **1e-12**, 1e-13), $\gamma$(0.001, 0.02, **0.05**, 0.1) | $\lambda_{ewc}$(1, 2, **5**), $\lambda_{af}$(1e-5, 1e-6, 1e-7, **1e-8**, 1e-9), $\gamma$(0.0001, **0.001**, 0.01, 0.1) | $\lambda_{ewc}$(40, 80, 160, **320**), $\lambda_{af}$(1e-3, **1e-4**, 1e-5, 1e-6), $\gamma$(0.0001, **0.001**, 0.01, 0.1) |
| CAF-MAS | $\lambda_{ref}$(4, 8, 16, **32**), $\lambda_{af}$(1e-10, 1e-11, **1e-12**, 1e-13), $\gamma$(0.05) | $\lambda_{ref}$(0.01, 0.02, **0.05**), $\lambda_{af}$(1e-6, 1e-7, 1e-8, **1e-9**), $\gamma$(0.0001, **0.001**, 0.01, 0.1) | $\lambda_{ref}$(**0.5**, 1), $\lambda_{af}$(1e-5, 1e-6, 1e-7, **1e-8**), $\gamma$(0.0001, **0.001**, 0.01, 0.1) |

# 3 Comparison with Preliminary Version

The ultimate goal of our research is to develop a continual learning model applicable to both artificial intelligence (AI) and biological intelligence (BI). Based on our preliminary versions of active forgetting and multi-learner architecture (presented in NeurIPS21 [18] and ECCV22 [19], respectively), the current manuscript is able to present a synergistic model of the *Drosophila* $\gamma$MB system for continual learning, with substantial extensions corresponding to both AI and BI:

**AI Aspect**: Technical improvement and adaptive cooperation of individual components.

• Our preliminary version of active forgetting [18] is basically AF-2, which can explicitly balance the new and old task solutions but results in more than twice the resource overhead, and provides mainly empirical evidence to support its advantages. In the current manuscript, we derive an efficient version from the Bayesian learning framework, i.e., AF-1, which is statistically equivalent to AF-2 but avoids additional resource overhead. We further provide theoretical evidence for the function of active forgetting from the perspective of generalization bound (Methods Proposition 2), applicable to both the SCL and MCL.

• Our preliminary version of the multiple-learner architecture [19] focuses on demonstrating an interesting finding that a specific implementation of multiple learners can improve continual learning performance. In the current manuscript, beyond a few architectural modifications for biological plausibility, we have extensively investigated the key factor of the neuro-inspired MCL, i.e., the diversity of learners' expertise, which is greatly affected by the innate randomness such as initialization and dropout. Besides, a large body of our ECCV paper is to provide a uniform generalization bound for continual learning performance. In the current manuscript, this serves as a broader objective to guide computational modeling of neurological adaptive mechanisms, and we further derive a tightened version of the generalization bound with recent advances in PAC-Bayes theory (Methods Proposition 2).

• The most interesting discovery is the synergistic cooperation of individual components, formulated respectively from the perspectives of Bayesian learning and neuroscience. In particular, the proposed two modulations (i.e., AF-1 and AF-S) can faithfully regulate the target distribution of each learner and thus properly coordinate the diversity of learners' expertise (Fig. 3). The full version of our approach can effectively address the identified objective of continual learning (Fig. 4, Fig. 6d,e, Supplementary Fig. 7), exhibiting outstanding performance and generality across a wide range of continual learning benchmarks (Fig. 5, Fig. 6).

**BI Aspect**: Extensive and in-depth biological plausibility.

• Thanks to the above efforts, the computational model in the current manuscript has

incorporated adaptive mechanisms of the *Drosophila* $\gamma$MB system for continual learning in a *synergistic* fashion at both *functional* and *architectural* levels. The strong computational evidence and biological plausibility enable us to validate and explain the corresponding biological processes, such as the biological significance and underlying mechanisms of active forgetting, as well as the adaptive modulations of forgetting rates and learning rules in multiple parallel compartments. These results are expected to facilitate subsequent interdisciplinary explorations.

# Supplementary Figures

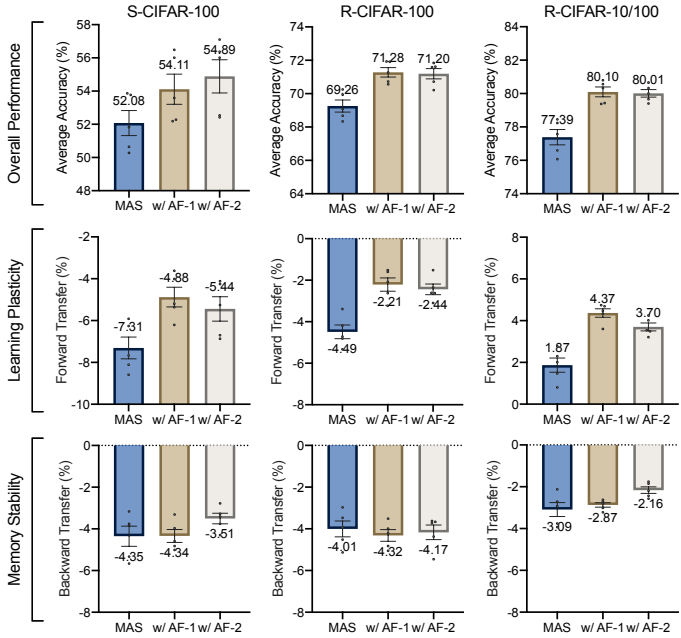

**Supplementary Fig. 1  Extended results of synaptic regularization.** Using MAS [1] as the baseline approach, the proposed active forgetting (AF-1 and AF-2) exhibits similar benefits in terms of overall performance, learning plasticity, and memory stability. All results are averaged over 5 runs with different random seeds and task orders. The error bars represent the standard error of the mean.

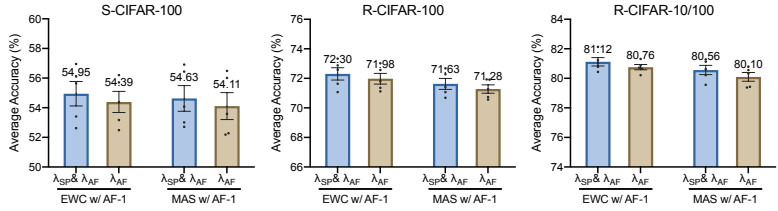

**Supplementary Fig. 2  Selection of hyperparameters for AF-1.** Performing a grid search of $\lambda_{SP}$ and $\lambda_{AF}$ together can slightly outperform that of $\lambda_{AF}$ only. All results are averaged over 5 runs with different random seeds and task orders. The error bars represent the standard error of the mean.

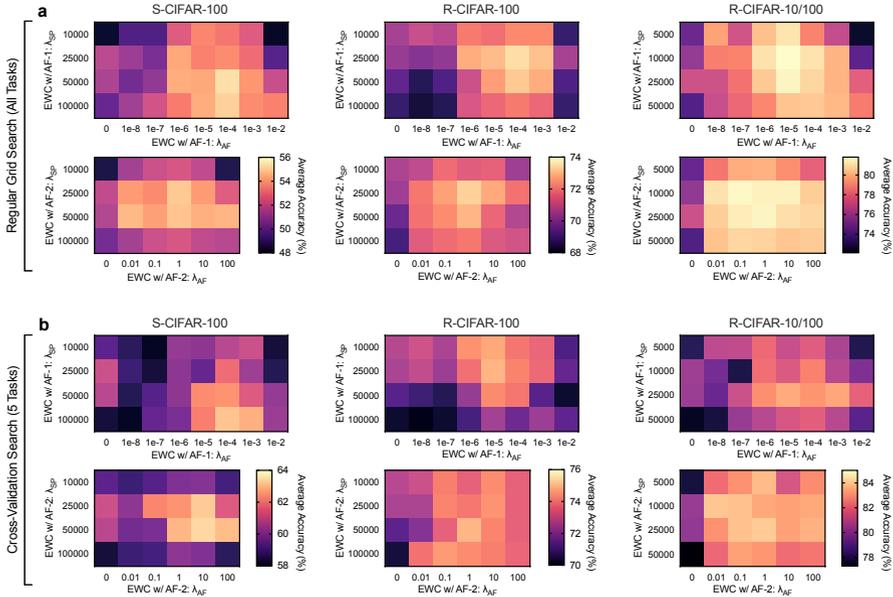

**Supplementary Fig. 3  Grid search of hyperparameters. a**, The grid search of hyperparameters is performed over the entire task sequence, which is a commonly-used protocol for continual learning papers. **b**, We follow the idea of cross-validation [4] and report the performance of different hyperparameters over a short sequence of 5 randomly-selected tasks, which is more practical in application. It can be seen that the good regions of cross-validation search are basically included in the counterpart of regular grid search.

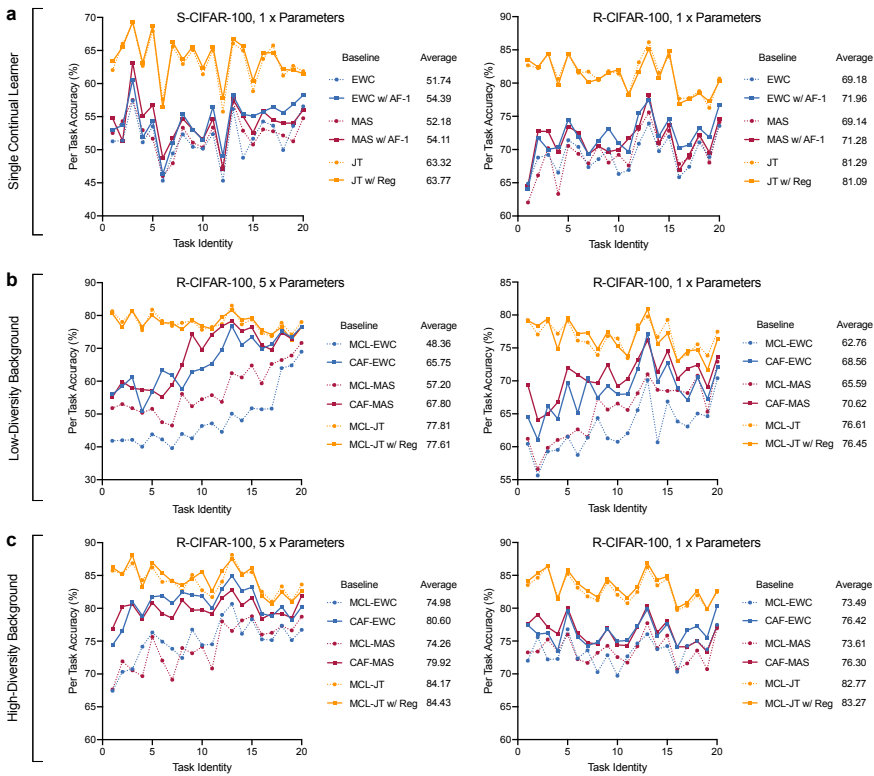

**Supplementary Fig. 4  Extended results of per-task performance.** The performance of each task is presented after continual learning or joint training (JT). We evaluate the effect of various factors, such as baseline approach, continual learning benchmark, network width and background diversity. The regularization used by our approach is also implemented in JT accordingly (refer to as "Reg"), with the same hyperparameters for fair comparison. It can be clearly seen that the performance improvements of ours are specific to continual learning. All results are averaged over 5 runs with different random seeds and task orders.

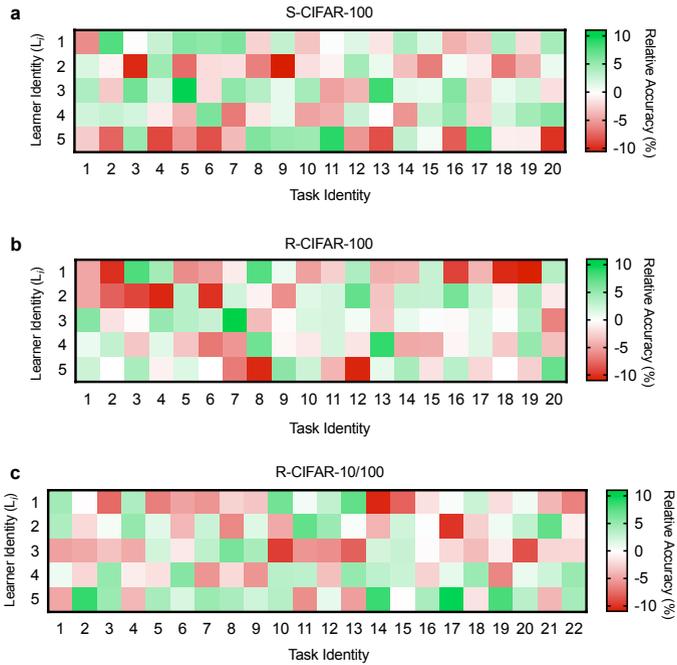

**Supplementary Fig. 5 Extended results of learners' expertise across tasks.** After continual learning of all tasks, we evaluate the relative accuracy of each learner across tasks, calculated as the performance of each learner minus the average performance of all learners. Here we use EWC [10] as the baseline approach for continual learning.

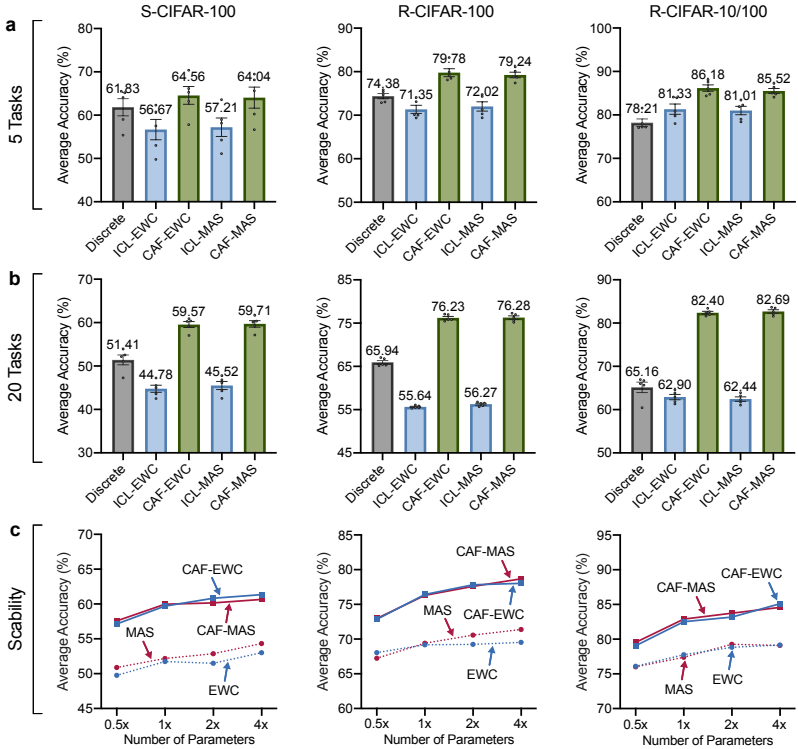

**Supplementary Fig. 6  Extended results of architecture evaluation. a, b,** We consider two multi-learner baselines to evaluate the architectural advantages of the neuro-inspired MCL under similar parameter budgets. Discrete: using a separate learner for each task. Independent Continual Learners (ICL): averaging the predictions of five independently-trained continual learners. **c,** The scalability of our approach is further evaluated by adjusting the network width of the MCL and the SCL under different parameter budgets. All results are averaged over 5 runs with different random seeds and task orders. The error bars represent the standard error of the mean.

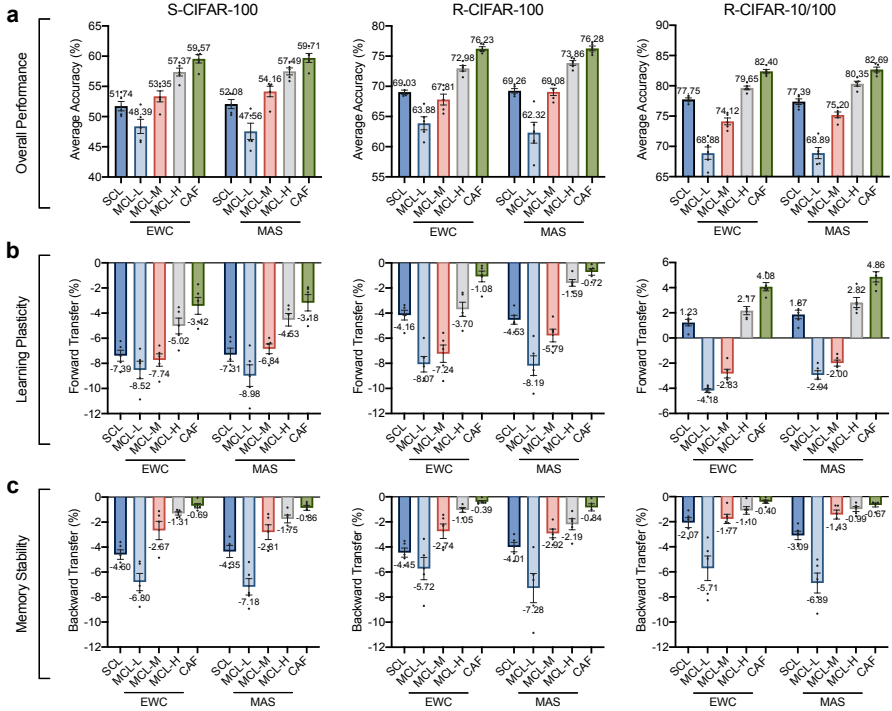

**Supplementary Fig. 7 Extended results of learner diversity and knowledge transfer.** We present a comprehensive comparison of the SCL and the $\gamma$MB-like architecture of MCL with different degrees of innate diversity. MCL-H (high-diversity): different random initialization with dropout. MCL-M (medium-diversity): identical random initialization with dropout. MCL-L (low-diversity): identical random initialization without dropout. **a**, Average accuracy for overall performance of all tasks. **b**, Forward transfer for learning plasticity of new tasks. **c**, Backward transfer for memory stability of old tasks. All results are averaged over 5 runs with different random seeds and task orders. The error bars represent the standard error of the mean. The network widths of MCL and CAF are adjusted accordingly to keep the total amount of parameters comparable to that of the SCL.



\end{document}